\documentclass[10pt,twocolumn,letterpaper]{article}

\usepackage[pagenumbers]{iccv} 

\usepackage{multirow}
\usepackage{xcolor}         
\usepackage{amsmath, amssymb, amsthm} 

\usepackage{bm}
\usepackage{graphicx}
\usepackage{svg}
\usepackage{multirow}
\usepackage{colortbl}

\usepackage{wrapfig}
\usepackage{pifont}
\usepackage{subcaption}
\usepackage{arydshln}
\usepackage{siunitx}
\usepackage{enumitem}

\newcommand{\cmark}{\ding{51}}%
\newcommand{\xmark}{\ding{55}}%

\usepackage{amsmath,amsfonts,bm}
\usepackage{xspace}

\providecommand{\eg}{\textit{e.g.}\@\xspace}
\providecommand{\ie}{\textit{i.e.}\@\xspace}

\def\eqref#1{equation~\ref{#1}}

\def\1{\bm{1}}

\def\rmX{{\mathbf{X}}}

\def\rmZ{{\mathbf{Z}}}

\DeclareMathAlphabet{\mathsfit}{\encodingdefault}{\sfdefault}{m}{sl}
\SetMathAlphabet{\mathsfit}{bold}{\encodingdefault}{\sfdefault}{bx}{n}

\definecolor{iccvblue}{rgb}{0.21,0.49,0.74}

\definecolor{ourscolor}{HTML}{DAE4F6}
\newcolumntype{a}{>{\columncolor{ourscolor}}c}

\usepackage[pagebackref,breaklinks,colorlinks,allcolors=iccvblue]{hyperref}

\def\paperID{8366} 
\def\confName{ICCV}
\def\confYear{2025}

\title{Harmonizing Visual Representations for\\ Unified Multimodal Understanding and Generation}

\author{
\centerline{
Size Wu\textsuperscript{\rm 1}\qquad
Wenwei Zhang\textsuperscript{\rm 2}\qquad
Lumin Xu\textsuperscript{\rm 3}\qquad
Sheng Jin\textsuperscript{\rm 4}\qquad
Zhonghua Wu\textsuperscript{\rm 5}
} \\
\centerline{
Qingyi Tao\textsuperscript{\rm 5}\qquad
Wentao Liu\textsuperscript{\rm 4}\qquad
Wei Li\textsuperscript{\rm 1}\qquad
Chen Change Loy\textsuperscript{\rm 1}
} \\
\centerline{
\textsuperscript{\rm 1} S-Lab, Nanyang Technological University \qquad
\textsuperscript{\rm 2} Shanghai AI Laboratory
Research
}\\
\centerline{
\textsuperscript{\rm 3} The Chinese University of Hong Kong 
 \quad 
\textsuperscript{\rm 4} SenseTime Research and Tetras.AI \quad
\textsuperscript{\rm 5} SenseTime Research
}\\
\centerline{
\url{size001@e.ntu.edu.sg} \qquad \url{{wei.l,ccloy}@ntu.edu.sg}
}\\
\vspace{5pt}
}

\begin{document}

\twocolumn[{%
\renewcommand\twocolumn[1][]{#1}%
\maketitle
\vspace{-20pt}
\centering
\includegraphics[width=1.0\textwidth]{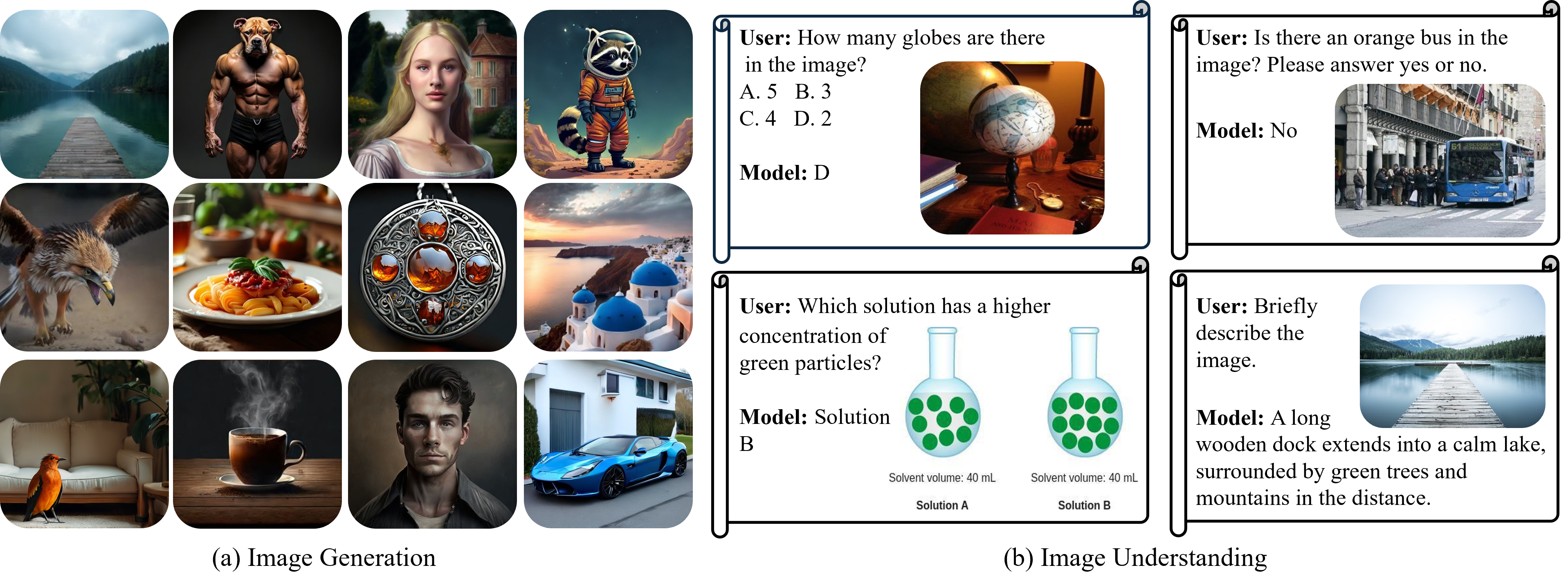}
\vspace{-20pt}
\captionof{figure}{An overview of image generation and understanding examples. All results are obtained by the proposed Harmon-1.5B, which uses a shared visual encoder for both tasks.}
\label{fig:flmm_teaser1}
\vspace{12pt}
}]

\begin{abstract}
Unifying visual understanding and generation within a single multimodal framework remains a significant challenge, as the two inherently heterogeneous tasks require representations at different levels of granularity.
Current approaches that utilize vector quantization (VQ) or variational autoencoders (VAE) for unified visual representation prioritize intrinsic imagery features over semantics, compromising understanding performance. In this work, we take inspiration from masked image modelling (MIM) that learns rich semantics via a mask-and-reconstruct pre-training and its successful extension to masked autoregressive (MAR) image generation. A preliminary study on the MAR encoder's representation reveals exceptional linear probing accuracy and precise feature response to visual concepts, which indicates MAR's potential for visual understanding tasks beyond its original generation role. Based on these insights, we present \emph{Harmon}, a unified autoregressive framework that harmonizes understanding and generation tasks with a shared MAR encoder. Through a three-stage training procedure that progressively optimizes understanding and generation capabilities, Harmon achieves state-of-the-art image generation results on the GenEval, MJHQ30K and WISE benchmarks while matching the performance of methods with dedicated semantic encoders (e.g., Janus) on image understanding benchmarks. Our code and models will be available at \url{https://github.com/wusize/Harmon}.

\end{abstract}    
\section{Introduction}
\label{sec:intro}

``\emph{What I cannot create, I do not understand.}''

~~~~~~~~~~~~~~~~~~~~~~~~~~~~~~~~~~~~~~~~~~~~~~~~~~~~~~~~~~~~~---~Richard Feynman
\vspace{0.1cm}

Generating images from textual descriptions and understanding images through targeted questions represent the forefront of AI research. With significant progress in both tasks, a plethora of text-to-image generation models~\cite{2022LDM, 2023Pixelartalpha, 2023SDXL, 2024lumina, 2024pixartsigma, ramesh2021zero, ramesh2022hierarchical, dalle3} and multimodal understanding LLMs~\cite{zhu2023minigpt, liu2024visual, liu2024improved, li2024llava, zang2025contextual, instructblip, yang2024qwen2, chen2024internvl, chen2024far,wu2024f} have been developed, driven by advancements in model architectures and computation scaling. To push the envelope even further, unifying these two tasks into a single and cohesive framework that handles both understanding and generation is highly desirable for next-generation multimodal intelligence.

Early attempts~\cite{sun2023generative, sun2024generative, tong2024metamorph, wang2024illume} have combined state-of-the-art diffusion models and multimodal LLMs by prompting the diffusion generators with LLM embeddings. These approaches are constrained by weak integration between image generation and text sequence modeling, resulting in suboptimal performance in instruction-based generation~\cite{ghosh2024geneval}.
More recently, a tighter coupling of generation and understanding has been achieved through either next-token prediction approaches~\cite{liu2024world, team2024chameleon, wu2024vila, wang2024emu3} or denoising diffusion frameworks~\cite{zhou2024transfusion, xie2024show, zhao2024monoformer, xiao2024omnigen}, where visual representations for both tasks are jointly modeled.

\begin{figure}[t]
  \centering
\hspace*{-2pt} 
\includegraphics[width=0.47\textwidth]{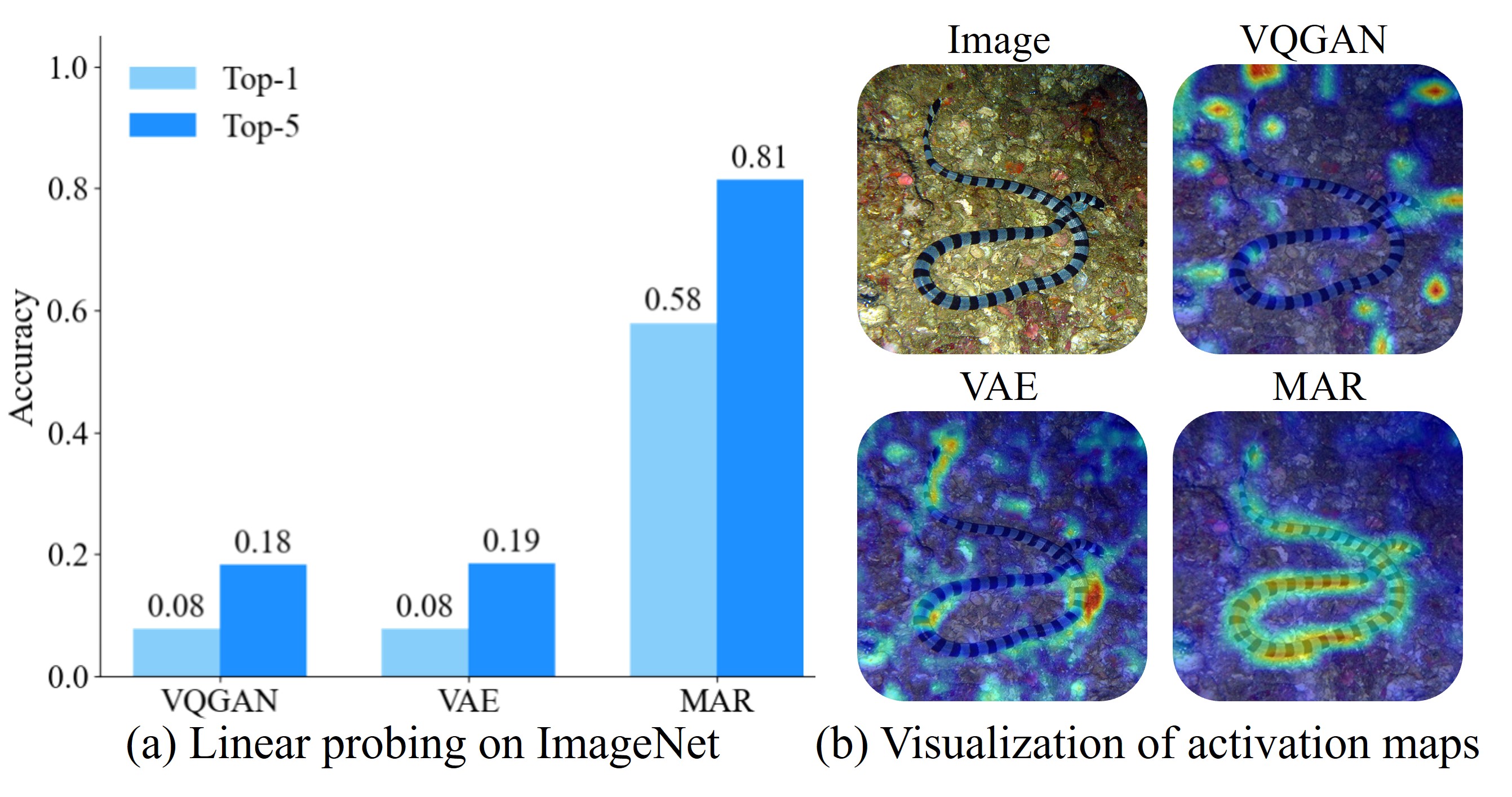}
  \vspace{-4pt}
  \caption{Linear probing and feature visualization (using GradCAM++~\cite{chattopadhay2018grad}) provide crucial insights into the suitability of representations learned by VQGAN, VAE, and MAR models for serving as the visual encoder in a unified multimodal framework.}
  \label{fig:pilot_study}
\vspace{-0.2cm}
\end{figure}

Visual understanding and generation are inherently heterogeneous tasks that require representations at different levels of granularity. In particular, visual understanding demands coarse-grained, high-level representations, whilst generation prefers fine-grained, intrinsic imagery features. 
Consequently, following standard practices in visual generation~\cite{2022LDM, sun2024autoregressive}, most approaches employ either a vector quantization (VQ) model or a variational autoencoder (VAE) to compress visual inputs into discrete tokens or continuous latents.
These encoders are primarily pre-trained to preserve intrinsic image features for pixel-level reconstruction rather than capturing visual semantics, resulting in limited image understanding capabilities. Therefore, to obtain a \textit{harmonious} representation with shared visual encoding for unified multimodal understanding and generation, a more integrated and moderate approach is warranted.


We take inspiration from masked image modeling (MIM)~\cite{he2022masked,bao2021beit} that can effectively encourage the model to develop a richer semantic representation of visual data via a simple \textit{mask-and-reconstruct} pretext task. This \textit{learning by generating} philosophy can learn balanced representations capable of both generation and understanding, making it a promising approach to reconcile these heterogeneous tasks in shared encoding. Moreover, the recent success of masked autoregressive (MAR)~\cite{li2024autoregressive, fan2024fluid} modeling that extends MIM to autoregressive image generation further substantiates the potential of MIM-based paradigms in this realm. It aligns with the de facto autoregressive prediction paradigm in MLLMs, making it more suitable for integrated synergy between generation and understanding. 

Driven by this revelation, we conduct preliminary studies involving linear probing and feature visualization to measure the representation power of widely used VQGAN and VAE encoders in generative models and the MAR encoder, which is shown in Figure~\ref{fig:pilot_study}. Our findings reveal that VQGAN and VAE encoders struggle to capture high-level semantics, such as object concepts and attributes, which are crucial for effective visual understanding. In contrast, the MIM-trained MAR encoder achieves competitive linear probing accuracy and produces precise feature map activations. This verifies our hypothesis that a MIM-based approach can yield more robust semantic representations, which are vital for unifying the tasks of generation and understanding.

\begin{figure}[t]
  \centering
\hspace*{-2pt} 
\includegraphics[width=0.47\textwidth]{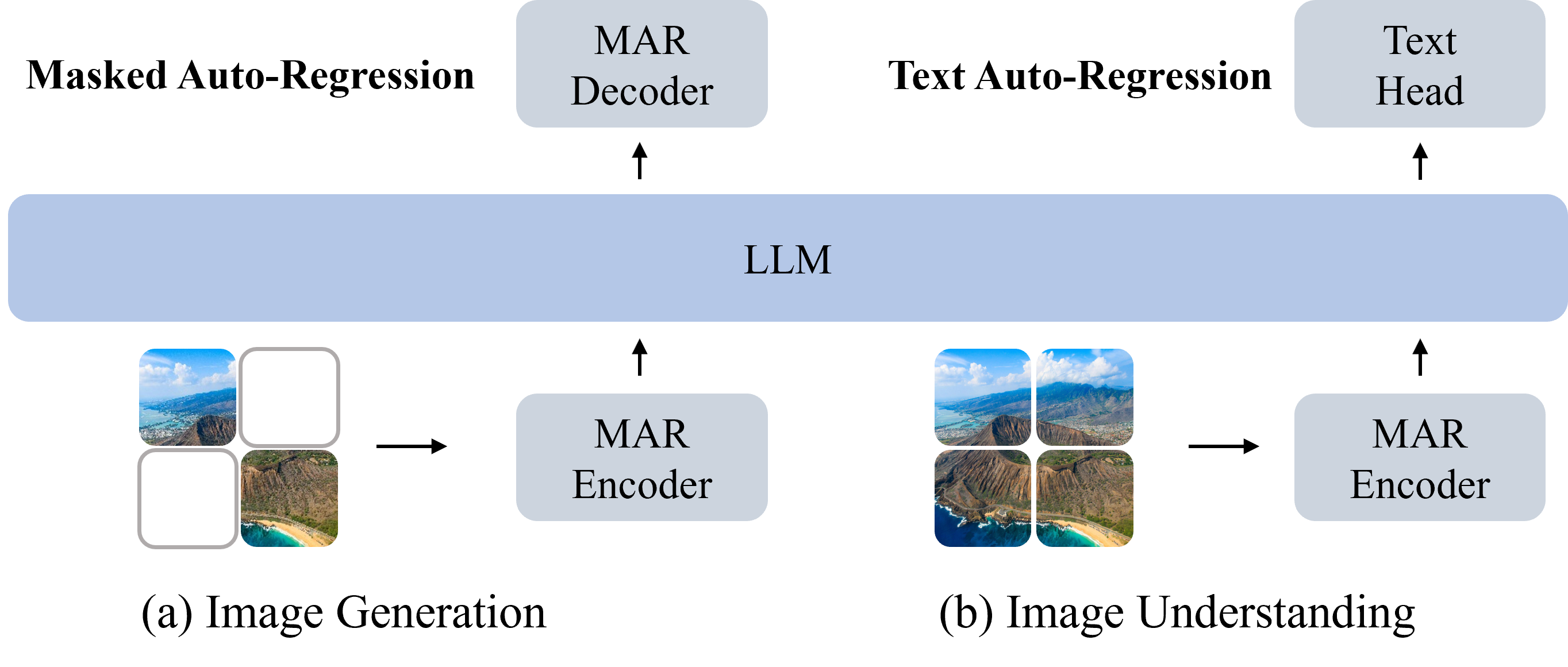}
  \vspace{-4pt}
  \caption{We propose Harmon, a unified framework with a shared MAR encoder for image generation and understanding tasks.}
  \label{fig:teaser2_simple}
\vspace{-0.2cm}
\end{figure}

Building on these insights, in this paper, we present \textit{Harmon}, a unified autoregressive model for multimodal generation and understanding as shown in Figure~\ref{fig:teaser2_simple}.
To harmonize visual representations for two tasks in shared encoding, we propose a unified visual encoder (MAR-based) that captures both coarse semantics and fine-grained features and promotes consistent semantic grounding for both tasks. In \textit{Harmon}, images are generated in a masked autoregressive manner while texts are regressed by next-token prediction.  Furthermore, to align this visual encoder with the LLM language space, we design a three-stage training pipeline to gradually develop a unified multimodal generation and understanding model. Thanks to the versatility of the MIM paradigm, the proposed MAR-based encoder can be thoroughly optimized for both generation and understanding in each training stage.

We comprehensively assess \textit{Harmon} on multimodal generation and understanding benchmarks. For text-to-image generation, it surpasses all unified methods with similar model scales on the GenEval benchmark~\cite{ghosh2024geneval}, which measures the alignment between generated images and user instructions. \textit{Harmon} also beats all unified models on the WISE benchmark~\cite{niu2025wise} that necessitates world knowledge comprehension, \textit{Harmon} produces images with higher visual quality, achieving state-of-the-art performance on the MJHQ30K benchmark~\cite{li2024playground}. For image understanding benchmarks, \textit{Harmon} performs competitively with methods adopting a separate semantic encoder for visual understanding (\ie, Janus~\cite{wu2024janus} and Janus-Pro~\cite{chen2025janus}) while surpassing methods that use VQ or VAE encoders by a large margin. Moreover, the synergy between the two tasks via a shared representation is also observed in our experiment, where image generation is improved by co-training with image understanding.

\section{Related Work}
\label{sec:related_work}

\noindent\textbf{Multimodal Understanding.} By injecting visual signals from a semantic encoder to an LLM, multimodal LLMs (MLLMs)~\cite{zhu2023minigpt, instructblip, liu2024visual, liu2024improved, li2024llava, lu2024deepseek, chen2024far, wang2024qwen2} demonstrate exceptional general and fine-grained image understanding capabilities. Such semantic encoders, including CLIP~\cite{radford2021learning} and SigLIP~\cite{zhai2023sigmoid}, are usually pre-trained via contrastive vision-language aligning, a process where comprehensive visual concepts are learned. Despite their powerful visual understanding capabilities, these MLLMs are limited to image-conditioned question-answering without the ability to produce visual outputs. 

\noindent\textbf{Image Generation.} Diffusion models~\cite{ramesh2021zero, 2022LDM, ramesh2022hierarchical, 2023SDXL, 2024pixartsigma, dalle3, peebles2023scalable, 2024hunyuandit} have become the dominant paradigm for image generation, producing high-quality visual contents conditioned on class labels or language descriptions. Typically, the diffusion process is instantiated in the latent space of VAE~\cite{van2017neural, kingma2013auto} for efficient training and inference. With the advancement of vision transformers, remarkable progress has also been made in autoregressive generation~\cite{esser2021taming,sun2024autoregressive, yu2023language, luo2024open} and masked generative models~\cite{chang2022maskgit, chang2023muse, bai2024meissonic}. In these frameworks, images are usually compressed into discrete tokens using a VQ model, \eg, VQGAN~\cite{esser2021taming}. More recently, the MAR~\cite{li2024autoregressive} paradigm is introduced, following an encoder-decoder architecture and leveraging masked image modelling~\cite{he2022masked} for image generation.

\noindent\textbf{Unified Visual Generation \& Understanding.} Pioneering works~\cite{sun2023generative, sun2024generative, tong2024metamorph, wang2024illume} usually combine state-of-the-art diffusion models and multimodal LLMs, prompting the diffusion generators with LLM embeddings. Such paradigm is limited by the lack of deep interaction between image generation and text sequence modeling, exhibiting sub-optimal performance in instruction-based generation~\cite{ghosh2024geneval}. Besides, a few studies unify generation and understanding tasks as next token prediction~\cite{liu2024world, team2024chameleon, wu2024vila, wang2024emu3}, where a VQGAN~\cite{esser2021taming} compresses images into discrete tokens. There are also studies~\cite{zhou2024transfusion, xie2024show, zhao2024monoformer, xiao2024omnigen} that regard LLMs as diffusion backbones, de-noising corrupted VAE latents for image generation but predicting next tokens for text regression. These two frameworks encode images using either VQGAN or VAE, which is a compression model by design, leading to diminution in visual understanding capability.

To address the above dilemma, ViLA-U~\cite{wu2024vila} builds a unified vision tower by learning vector quantization on visual foundation models with joint contrastive text alignment and image reconstruction objectives. However, it struggles to balance the intricate interplay between semantic alignment and pixel-level fidelity. There are also studies that disentangle the forward pathways of understanding and generation. LlamaFusion~\cite{shi2024llamafusion} learns extra model weights that are only activated for vision signals. Likewise, Janus~\cite{wu2024janus, chen2025janus} employs separate visual encoders for the two tasks, \ie, VQGAN~\cite{esser2021taming} for generation and SigLIP~\cite{zhai2023sigmoid} for understanding. Despite achieving state-of-the-art performances, this ad-hoc design neglects the potential of a unified representation that fosters deeper cross-modal synergy across both tasks. In contrast, we find that the MIM-trained MAR encoder naturally learns a representation well-suited for both generation and understanding. Based on this insight, we introduce Harmon, a unified framework that employs a shared encoder for both tasks.

\begin{figure*}[t]
  \centering
  \includegraphics[width=0.98\textwidth]{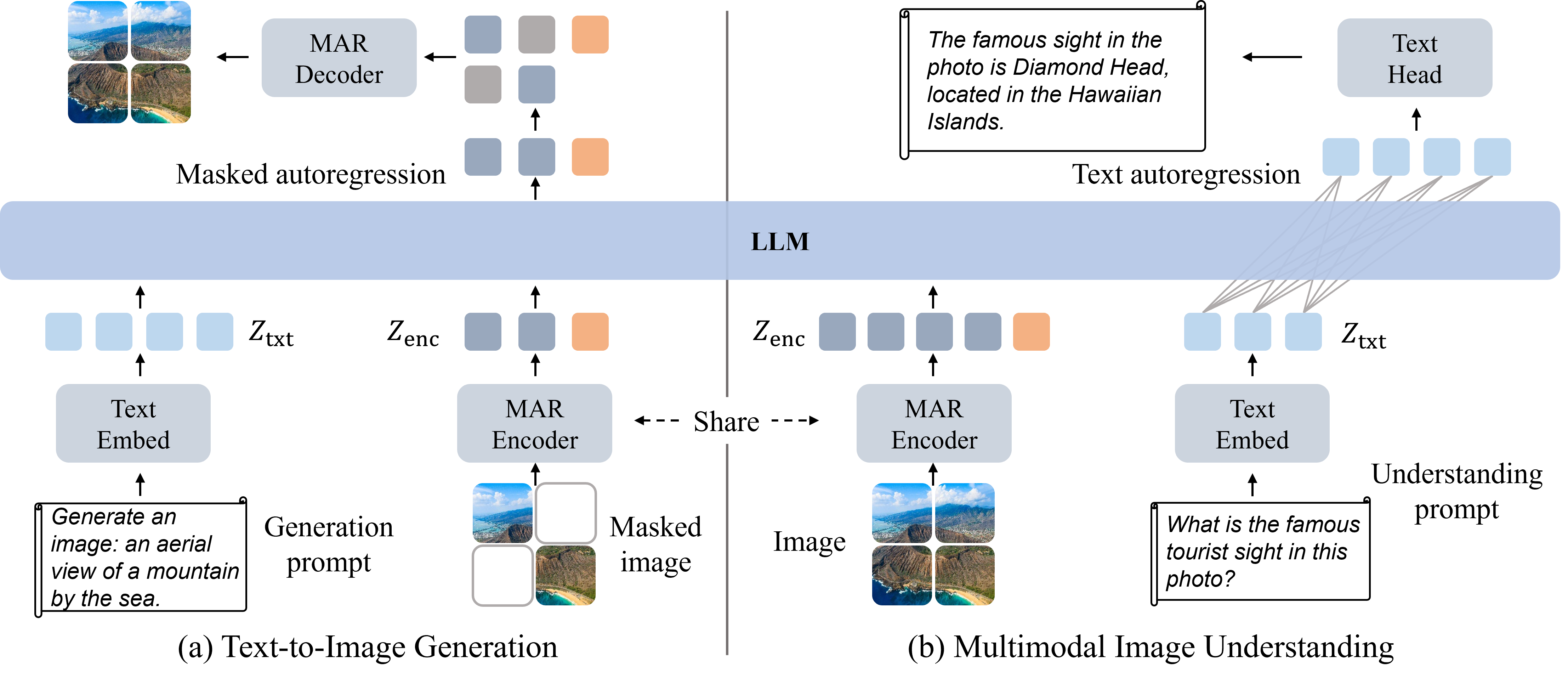}
\vspace{-4pt}
  \caption{The overall framework of Harmon. (a) Image generation is performed in the masked autoregressive manner. (b) Image understanding is formulated as image-conditioned text autoregression. The MAR encoder is shared by both tasks.  {\includegraphics[height=1.5ex]{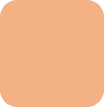}} represents the buffer embedding and  {\includegraphics[height=1.5ex]{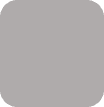}} represents the mask embedding.
  }
  \label{fig:method}
\end{figure*}

\section{Method}

In this section, we introduce Harmon, a unified autoregressive framework comprising an MAR encoder, an LLM, and an MAR decoder as shown in Figure~\ref{fig:method}. For brevity, we omit the connection layers between the LLM and the MAR encoder/decoder.

\subsection{Preliminaries}
\label{sec:preliminary}
\noindent\textbf{Masked Autoregressive Model.} MAR~\cite{li2024autoregressive} is a masked autoregressive model for image generation. Our preliminary results in Figure~\ref{fig:pilot_study} suggest that the MAR encoder actually captures visual concepts and semantics during its generative pre-training, which are critical for visual understanding. This makes it a viable choice as an encoder to handle both visual generation and understanding within a single, unified framework beyond its original role.

In an MAR forward pass where $m$ patches are masked from an image $\rmX_{\text{img}}$ with height $h$ and width $w$, the encoder $f_{\text{enc}}$ takes the $hw - m$ seen image patches $\rmX_{\text{seen}}$ and $n$ buffer embeddings $\rmX_{\text{buffer}}$ as input for feature extraction 
\begin{equation}
\label{eq:mim_enc}
\rmZ_{\text{enc}} = f_{\text{enc}}(\rmX_{\text{seen}}, \rmX_{\text{buffer}}).
\end{equation}
Then $\rmZ_{\text{enc}}$ outputted by the encoder and a set of mask embeddings $\rmZ_{\text{mask}}$ are fed into the decoder $f_{\text{enc}}$, which predicts the $m$ masked image patches
\begin{equation}
\rmX_{\text{mask}} = f_{\text{dec}}(\rmZ_{\text{mask}}, \rmZ_{\text{enc}}).
\end{equation}

\noindent\textbf{Large Language Model.} A modern large language model (LLM) typically follows GPT~\cite{radford2018improving, radford2019language, brown2020language} to adopt a decoder-only architecture with causal attention. In a forward pass, a sequence of text tokens are first mapped into text embeddings and then fed into the LLM $f_{\text{LLM}}$. Each token embedding interacts with preceding ones in the LLM's causal attentions and predicts the next token by a text head on top of the LLM.

\subsection{Text-to-Image Generation}
Image generation in Harmon works in a masked autoregressive manner as illustrated by Figure~\ref{fig:method}(a).

\noindent\textbf{Task Prompt.} For text-to-image generation, we use the following prompt template to format user instruction: ``\texttt{User: Generate an image <caption>\textbackslash n Assistant:}''. \texttt{<caption>} represents the image description. The text prompt is tokenized and mapped to text embeddings $\rmZ_{\text{txt}}$ and then passed to the LLM to guide image generation.  During training, \texttt{<caption>} is randomly set to empty for 10\% data samples to enable classifier-free guidance (CFG)~\cite{ho2022classifier} in inference.

\noindent\textbf{Encoder to LLM.} As shown in Eq.~\ref{eq:mim_enc} and Figure~\ref{fig:method}(a), the output of the MAR encoder $\rmZ_{\text{enc}}$ includes image embeddings of seen patches and buffer embeddings.
Instead of directly feeding $\rmZ_{\text{enc}}$ to the MAR decoder for reconstruction, we pass $\rmZ_{\text{enc}}$ to the LLM following the prompt text embeddings $\rmZ_{\text{txt}}$ for multimodal integration.

\noindent\textbf{LLM to Decoder.} The output produced by the LLM consists of the processed prompt embeddings and MAR encoder outputs. We extract the updated encoder outputs $\rmZ'_{\text{enc}}$, and feed them to the MAR decoder $f_{\text{dec}}$ together with the $m$ mask embeddings $\rmZ_{\text{mask}}$. Then, $f_{\text{dec}}$ predicts the masked image patches
\begin{equation}
\rmX_{\text{mask}} = f_{\text{dec}}(\rmZ_{\text{mask}}, \rmZ'_{\text{enc}}).
\end{equation}
Since $\rmZ'_{\text{enc}}$ has interacted with text embeddings in the LLM, the prediction is conditioned on the language prompts.

\begin{table*}[h]
  \centering
\caption{\textbf{Detailed hyperparameters in training}. Data ratio refers to the ratio of text data, image-to-text data, and text-to-image data.}
\scalebox{0.9}
{\begin{tabular}{cccc}
\toprule
Setting & Stage I & Stage II &  Stage III \\
\midrule
 & MAR Decoder: $10^{-4}$ & MAR Decoder: $ 10^{-4}$ & MAR Decoder: $ 2\times10^{-5}$ \\
\multirow{-2}[0]{*}{LR.} & MAR Encoder: $ 10^{-5}$  & LLM \& MAR Encoder: $ 10^{-5}$ & LLM \& MAR Encoder: $ 2\times10^{-6}$ \\
Image Size & 256 & 256 & 512 \\
Batch Size & 4096 & 4096 & 1024 \\
Data Ratio & 0:1:2 & 1:3:8 & 1:3:16  \\

Training Step & 50,000 & 50,000 & 50,000 \\

\bottomrule
\end{tabular}}
\label{tab:hyperparameters}
\end{table*}

\noindent\textbf{Training Objective.} We follow MAR~\cite{li2024autoregressive} to reparameterize the generation of masked patches as a reverse diffusion process~\cite{ho2020denoising}. Therefore, the training loss for predicting a masked patch $x_{\text{mask}} \in \rmX_{\text{mask}}$ is 
\begin{equation}
\mathcal{L}(x_{\text{mask}}, x_{\text{gt}}) = \mathbb{E}_{\varepsilon, t} \left[ \left\| \varepsilon - \varepsilon_\theta(x_t | t,x_{\text{mask}}) \right\|^2 \right].
\label{eq:denoise}
\end{equation}
Here, $\varepsilon$ is a noise vector sampled from $\mathcal{N}( \mathbf{0}, \mathbf{I})$.
$x_t$ is a noise-corrupted vector from $x_{\text{gt}}$ and $t$ is a randomly sampled time step of the noise schedule.
The noise estimator $\varepsilon_\theta$ is a small MLP network defined in ~\cite{li2024autoregressive}.

\noindent\textbf{Inference.} Our masked autoregressive generation consists of $K$ forward passes, starting with the masked patch number $m_0=hw$, \ie, all image patches are masked and only the buffer embeddings are fed into the MAR encoder. We gradually reduce the mask ratio following a cosine curve: $m_k = hw\cdot\text{cos}(\frac{k}{2K}\pi)$. Here, $k$ means the $k$-th generation step. And the number of predicted image patches is $m_k - m_{k-1}$ in the $k$-th step. Since the attention in the LLM is causal, prompt embeddings are only passed to the LLM in the first iteration, leaving key-value caches for later steps. In addition, we also follow prior works~\cite{ li2024autoregressive, xie2024show, wang2024emu3} to employ the classifier-free guidance (CFG)~\cite{ho2022classifier}.

\subsection{Image Understanding}
As shown in Figure~\ref{fig:method}(b),
multimodal image understanding is formulated as question-answering conditioned on image contents, where the text autoregression is performed as next-token prediction.

\noindent\textbf{Task Prompt.} We use the following prompt template to format user instruction: ``\texttt{User:<image><question>\textbackslash n Assistant:}''. \texttt{<question>} represents the question related to the image and \texttt{<image>} is the output $\rmZ_{\text{enc}}$ from the MAR encoder $f_{\text{enc}}$.

\noindent\textbf{Visual Encoding.} Different from the text-to-image generation process, we pass all the image patches without any masking together with the buffer embeddings to the MAR encoder $f_{\text{enc}}$
\begin{equation}
\rmZ_{\text{enc}} = f_{\text{enc}}(\rmX_{\text{img}}, \rmX_{\text{buffer}}).
\end{equation}

\noindent\textbf{Answer Generation.} The encoder output $\rmZ_{\text{enc}}$ together with text embeddings in the prompt are fed to the LLM. The corresponding textual answer is generated by the LLM in an autoregressive manner, conditioned on the image and the question. A cross-entropy loss is applied to supervise the prediction of answer tokens during training.

\subsection{Training Recipe}
\label{subsec:training}
The training stages and the associated data sources play a pivotal role in unleashing the potential of the shared encoder in Harmon. We split the training of Harmon into three stages, progressively enhancing the model's generation and understanding abilities.  More information about the data sources is provided in the appendix. 

\noindent\textbf{Stage I: Vision-Language Alignment.} Since the MAR encoder/decoder and the LLM undergo unimodal pre-training on images and texts, respectively, it is necessary to align these two modalities on a large corpus of image-text pairs. Specifically, we use 22M images with dense captions from public datasets~\cite{li2024llava, singla2024pixels, li2024densefusion, li2024mini} for knowledge-rich training of the MAR encoder, enabling language-based image understanding in Harmon. To maintain the original generation capability of MAR, we employ ImageNet1K~\cite{deng2009imagenet} with 1.2M data samples for class-conditional generation, treating class names as image captions. In this stage, only the MAR encoder and decoder are trainable.

\noindent\textbf{Stage II: Comprehensive Multimodal Training.} After aligning the MAR encoder/decoder with the LLM, we utilize diverse question-answering data and text-to-image data to enhance generation and understanding abilities, with the LLM unlocked. Specifically, we draw 20M samples from Infinity-MM~\cite{gu2024infinity} plus 5M dense caption samples from stage I for multimodal understanding. For text-to-image generation, we collect 50M images from public datasets~\cite{meyer2024public, madebyollin_megalith_10m, dclure_laion_aesthetics_12m_umap, singla2024pixels, li2024llava,li2024densefusion}. All of these images come with dense captions generated by LMMs~\cite{team2023gemini, chen2023sharegpt4v, wang2024qwen2}, which are re-written by an LLM~\cite{touvron2023llama2} for more concise prompts focusing on generation.

\noindent\textbf{Stage III: High-Quality Fine-tuning.} We adopt higher-quality data to fine-tune Harmon in stage III. For multimodal question-answering, we employ the smaller but more balanced instruction-tuning data from LLaVA-One-Vision~\cite{li2024llava}. For text-to-image generation, we filter the 50M images in stage II based on image sizes and aesthetic scores, with 20\% of them reserved. In addition, we include 6M synthetic images~\cite{pan2023journeydb, t2i2m}  to enhance visual quality. 
In stage III, we also increase the image resolution from 256 to 512 to strengthen both generation and understanding capabilities.

\begin{table*}[ht]
  \centering
\caption{Evaluation of image understanding capabilities on multimodal question-answering benchmarks. \textit{Und. Only} stands for models trained for image understanding only. $^\dagger$ means employing a separate semantic encoder.}
\scalebox{0.75}
{    \begin{tabular}{lllcccccccc}
        \toprule
        \textbf{Type} & \textbf{Model} & \textbf{Encoder} & \textbf{LLM Scale} & \textbf{POPE$ \uparrow$} & \textbf{MME-P$ \uparrow$} & \textbf{MME-C$ \uparrow$} & \textbf{MMB$ \uparrow$} & \textbf{SEED$ \uparrow$}  & \textbf{GQA$ \uparrow$} & \textbf{MMMU$ \uparrow$} \\
        \midrule
        \multirow{6}{*}{\textit{Und. Only}} &
        LLaVA-v$1.5$-Phi-$1.5$~\cite{xie2024show} & CLIP ViT-L~\cite{radford2021learning} & 1.3B & 84.1 & 1128 &- & - & -  & 56.5 & 30.7 \\
        & MobileVLM~\cite{chu2023mobilevlm} & CLIP ViT-L~\cite{radford2021learning}&1.4B & 84.5 & 1196 &- &53.2 & - &  56.1 & -\\
        & MobileVLM-V2~\cite{chu2024mobilevlm2} & CLIP ViT-L~\cite{radford2021learning}& 1.4B & 84.3 & 1303 &-& 57.7 & - &  59.3 & -\\

        & DeepSeekVL~\cite{lu2024deepseek} &SigLIP-Large~\cite{zhai2023sigmoid}& 1.3B & 88.3 & 1307 &225 & 64.6 & -& 59.3 & 33.8\\
        & \multirow{2}{*}{MiniGemini~\cite{li2024mini}}& CLIP ViT-L~\cite{radford2021learning}& \multirow{2}{*}{2B}& \multirow{2}{*}{83.9}& \multirow{2}{*}{1341} & \multirow{2}{*}{312} & \multirow{2}{*}{59.8} & \multirow{2}{*}{-}
        & \multirow{2}{*}{59.9} & \multirow{2}{*}{-} \\
        &&\&ConvNext-L~\cite{liu2022convnet}&\\
        \midrule
        \textit{Unified} 
        & ILLUME~\cite{wang2024illume} &OpenCLIP ViT-H~\cite{cherti2023reproducible}& 7B &  88.5 &  1445 &- & 65.1 &  72.9 &  - & 38.2  \\
        & TokenFlow-XL~\cite{qu2024tokenflow} &CLIP ViT-B(VQ)~\cite{qu2024tokenflow} & 13B & 86.8 & 1546 & - & 68.9 &  68.7 &  62.7 & 38.7   \\
        & LWM~\cite{liu2024world} &VQGAN~\cite{esser2021taming} & 7B & 75.2 & - & -&- & - &  44.8 & - \\
        & VILA-U~\cite{wu2024vila} &SigLIP-Large(VQ)~\cite{wu2024vila} & 7B & 85.8 & 1402& & - & 59.0 &  60.8 & - \\
        & Chameleon~\cite{team2024chameleon} &VQGAN~\cite{esser2021taming}& 7B & - & - & - &-& - &  - & 22.4 \\
 \cline{2-11}     
        & D-Dit~\cite{li2024dual} &VAE~\cite{esser2024scalingrectifiedflowtransformers} &2.0B & 84.0 & 1125 & - & -& - & \underline{59.2} & - \\
        & Show-o~\cite{xie2024show} & MAGVIT-v2~\cite{yu2023language}& 1.3B & 80.0 & 1097 & 248& 51.6 & 54.4 &  58.0 & 26.7 \\

        & Janus$^\dagger$~\cite{wu2024janus} &SigLIP-Large~\cite{zhai2023sigmoid} & 1.3B & \underline{87.0} & \underline{1338} &222 & \underline{69.4} & 63.7 &  59.1 & 30.5 \\
        & Janus-Pro$^\dagger$~\cite{chen2025janus} &SigLIP-Large~\cite{zhai2023sigmoid} & 1.5B & 86.2 & \textbf{1444}& \underline{268} & \textbf{75.5} & \textbf{68.3} & \textbf{59.3} & \underline{36.3} \\
        & \textbf{Harmon-0.5B} & MAR-B~\cite{li2024autoregressive}& 0.5B & 86.5 & 1148 & 260 & 59.8& 62.5 &56.3 &34.2 \\

        & \textbf{Harmon-1.5B} & MAR-H~\cite{li2024autoregressive}& 1.5B & \textbf{87.6} & 1155 & \textbf{321} & 65.5 & \underline{67.1} &58.9 & \textbf{38.9} \\

        \bottomrule
    \end{tabular}}
\label{tab:qa_benchmarks}
\end{table*}

\begin{table}[t]
  \centering
\caption{Text-to-image generation on MSCOCO-30K and MJHQ-30K. FID is used as the metric for both benchmarks. \textit{Gen. Only} stands for models trained for image generation only.}
\scalebox{0.8}
{    \begin{tabular}{llcc}
        \toprule
      \textbf{Type} & \textbf{Model} & \textbf{MSCOCO$\downarrow$} & \textbf{MJHQ$\downarrow$} \\
        \midrule
 \multirow{7}{*}{\textit{Gen. Only}} 
        &  DALL·E 2~\cite{ramesh2022hierarchical} & $10.39$ &- \\
        &  GigaGAN~\cite{kang2023scaling} & $9.09$ & - \\
        &  SD1.5~\cite{rombach2022high} & $9.62$ &-\\
        &  PixArt-$\alpha$~\cite{chen2023pixart} & 7.32& 9.85 \\
        & SDXL~\cite{2023SDXL} & 7.38&  8.76 \\
        & SD2.1~\cite{2022LDM} & - & 26.96 \\
        & LlamaGen~\cite{sun2024autoregressive} & - & 25.59 \\

    \midrule
 \multirow{8}{*}{\textit{Unified}} 
      & Show-o~\cite{xie2024show} & 9.24 & $15.18$ \\
       & LWM~\cite{liu2024world} & 12.68 & $17.77$ \\
      & VILA-U~\cite{wu2024vila} & - & $7.69$ \\
       & Janus~\cite{wu2024janus}& \underline{8.53}  & $10.10$ \\
       & Janus-Pro-1.5B~\cite{chen2025janus} & 16.08 & 9.53 \\
       & \textbf{Harmon-0.5B}  & 8.86 & \underline{6.08} \\
       & \textbf{Harmon-1.5B}  &\textbf{8.39} & \textbf{5.15} \\
        \bottomrule
    \end{tabular}}
\label{tab:fid_benchmarks}
\end{table}

\section{Experiments}
\label{sec:exp}

We conduct our experiments on two model variants, Harmon-0.5B and Harmon-1.5B. Harmon-0.5B is built on Qwen2.5-0.5B-Instruct~\cite{yang2024qwen2} with visual encoder and decoder from MAR-B~\cite{li2024autoregressive} while Harmon-1.5B is based on Qwen2.5-1.5B-Instruct~\cite{yang2024qwen2} and MAR-H~\cite{li2024autoregressive}. Image size is fixed at $256 \times 256$ in stage I and stage II, which is increased to $512 \times 512$ in stage III. All hyperparameters in the three-stage training are detailed in Table~\ref{tab:hyperparameters}. The whole training process costs $4$ days and $8$ days for Harmon-0.5B and Harmon-1.5B, respectively, with $32$ Nvidia A$100$ ($80$GB) GPUs. In inference, greedy search is employed for text generation in image understanding. For text-to-image generation, we use $K=64$ MAR forward passes and set CFG weight as 3.0.

\subsection{Comparisons with Existing Models}

\noindent \textbf{Image Understanding.}
We evaluate our model's image understanding capabilities on widely used multimodal question-answering benchmarks, including POPE~\cite{li2023evaluating}, MME~\cite{fu2023mme}, MMB~\cite{liu2023mmbench}, SEED \cite{li2023seed},  GQA~\cite{hudson2019gqa} and MMMU \cite{yue2024mmmu}. As shown in Table~\ref{tab:qa_benchmarks}, we compare Harmon with both unified models and understanding-only models (indicated as \textit{Und. Only}). Given the model size of Harmon, we mainly focus on multimodal models with LLM under 2B parameters. It is noteworthy that Harmon achieves performance on par with understanding-only models and Janus models, which employ a separate semantic encoder (\ie, SigLIP~\cite{zhai2023sigmoid}) for visual perception. Besides, Harmon outperforms prior unified models based on VAE/VQGAN encoders, such as LWM~\cite{liu2024world}, Chameleon~\cite{team2024chameleon}, D-DiT~\cite{li2024dual} and Show-O~\cite{xie2024show}.

\noindent\textbf{Text-to-Image Generation.} We assess both the visual quality and controllability of image generation. Specifically, we report Fréchet Inception Distance (FID) scores on the MSCOCO-$30$K \cite{chen2015microsoft} and MJHQ-$30$K \cite{li2024playground} benchmarks, measuring the similarity between the distributions of generated images and reference images. Besides, we evaluate our models on the GenEval \cite{ghosh2024geneval} benchmark, which examines how well the attributes of generated objects can be controlled by user instructions, including counting, position and colour attributes. As shown in Table~\ref{tab:fid_benchmarks}, Harmon outperforms all unified models on MJHQ-$30$K that gauges the aesthetic quality of generated images. For evaluation of controllability on GenEval, Harmon outperforms all unified and generation-only models as shown in Table~\ref{tab:geneval}. Notably, our smaller variant Harmon-0.5B is already comparable to Janus-Pro-1.5B, indicating the superior vision-language alignment brought by the shared visual encoder. Additionally, we assess Harmon's ability to understand complex semantic and world knowledge using the WISE benchmark~\cite{niu2025wise}, where implicit prompts like ``\texttt{Einstein's favorite musical instrument}'' are provided. As shown in Table~\ref{tab:wise}, Harmon archives the best performance among all unified models.

\begin{table*}[t]
  \centering
\caption{Evaluation of text-to-image generation on GenEval benchmark. \textit{Gen. Only} stands for models trained for image generation only.}
\scalebox{0.83}
{    \begin{tabular}{llcccccca}
        \toprule
        \textbf{Type} & \textbf{Method}  & \textbf{Single Obj.} & \textbf{Two Obj.} & \textbf{Counting} & \textbf{Colors} & \textbf{Position} & \textbf{Color Attri.} & \textbf{Overall$\uparrow$} \\
        \midrule
        \multirow{9}{*}{\textit{Gen. Only}} 
        & LlamaGen~\cite{sun2024autoregressive}  & 0.71 & 0.34 & 0.21 & 0.58 & 0.07 & 0.04 & 0.32 \\
        & LDM~\cite{rombach2022high} & 0.92 & 0.29 & 0.23 & 0.70 & 0.02 & 0.05 & 0.37 \\
        & SDv1.5~\cite{rombach2022high} &  0.97 & 0.38 & 0.35 & 0.76 & 0.04 & 0.06 & 0.43 \\
        & PixArt-$\alpha$~\cite{chen2023pixart} &  0.98 & 0.50 & 0.44 & 0.80 & 0.08 & 0.07 & 0.48 \\
        & SDv2.1~\cite{rombach2022high} &  0.98 & 0.51 & 0.44 & 0.85 & 0.07 & 0.17 & 0.50 \\
        & DALL-E 2~\cite{ramesh2022hierarchical}  & 0.94 & 0.66 & 0.49 & 0.77 & 0.10 & 0.19 & 0.52 \\
        & Emu3-Gen ~\cite{wang2024emu3}  & 0.98 & 0.71 & 0.34 & 0.81 & 0.17 & 0.21 & 0.54 \\
        & SDXL~\cite{2023SDXL} &  0.98 & 0.74 & 0.39 & 0.85 & 0.15 & 0.23 & 0.55 \\
        & DALL-E 3~\cite{dalle3}  & 0.96 & 0.87 & 0.47 & 0.83 & 0.43 & 0.45 & 0.67 \\
        & SD3-Medium~\cite{esser2024scalingrectifiedflowtransformers} & 0.99 & 0.94 & 0.72 & 0.89 & 0.33 & 0.60 & 0.74 \\
        \midrule
        \multirow{13}{*}{\textit{Unified}}

        & LWM~\cite{liu2024world} &  0.93 & 0.41 & 0.46 & 0.79 & 0.09 & 0.15 & 0.47 \\
        
        & SEED-X~\cite{ge2024seed}  & 0.97 & 0.58 & 0.26 & 0.80 & 0.19 & 0.14 & 0.49 \\
        & Show-o~\cite{xie2024show} &  0.95 & 0.52 & 0.49 & 0.82 & 0.11 & 0.28 & 0.53 \\
        & D-DiT~\cite{li2024dual} &  0.97 & 0.80 & 0.54 & 0.76 & 0.32 & 0.50 & 0.65 \\

        & Transfusion~\cite{zhou2024transfusion} & - & - & - & - & - & - & 0.63 \\
        & ILLUME~\cite{wang2024illume} &  0.99 & 0.86 & 0.45 & 0.71 & 0.39 & 0.28 & 0.61 \\
        & TokenFlow-XL~\cite{liu2024world} &  0.95 & 0.60 & 0.41 & 0.81 & 0.16 & 0.24 & 0.55 \\
        & OmniGen~\cite{xiao2024omnigen} & 0.99 & 0.86 & 0.64 & 0.85 &0.31 &0.55 & 0.70 \\
        & Chameleon~\cite{team2024chameleon} &  - & - & - & - & - & - & 0.39 \\
        & Janus~\cite{wu2024janus} & 0.97 & 0.68 & 0.30 & 0.84 & 0.46 & 0.42 & 0.61 \\
        & Janus-Pro-1.5B~\cite{chen2025janus} &  0.98 & 0.82 & 0.51 & 0.89 & 0.65 & 0.56 & \underline{0.73} \\
        & \textbf{Harmon-0.5B} &  0.99 & 0.80 & 0.57 & 0.87 & 0.55 & 0.48 & 0.71 \\
        & \textbf{Harmon-1.5B} &  0.99 & 0.86 & 0.66 & 0.85 & 0.74 & 0.48 & \textbf{0.76} \\
        \bottomrule
    \end{tabular}

    }
\label{tab:geneval}
\end{table*}

\begin{table*}[t]
  \centering
\caption{Evaluation of text-to-image generation on WISE benchmark. \textit{Gen. Only} stands for models trained for image generation only.}
\scalebox{0.88}
{    \begin{tabular}{llcccccca}
        \toprule
        \textbf{Type} & \textbf{Method}  & \textbf{Cultural} & \textbf{Time} & \textbf{Space} & \textbf{Biology} & \textbf{Physics} & \textbf{Chemistry} & \textbf{Overall$\uparrow$} \\
        \midrule
        \multirow{9}{*}{\textit{Gen. Only}} 
        & SDv1.5~\cite{rombach2022high} &  0.34 & 0.35& 0.32&0.28 &0.29 &0.21 &  0.32\\
        & SDv2.1~\cite{rombach2022high} &  0.30 & 0.38 &0.35 & 0.33 & 0.34&0.21 & 0.32 \\
        & Emu3-Gen ~\cite{wang2024emu3}  &0.34&0.45&0.48 &0.41 &0.45 &0.27 & 0.39 \\

        & FLUX.1-schnell~\cite{flux2024} &0.39  &0.44  &0.50 & 0.31&0.44  &0.26  & 0.40 \\

        & SD3-Medium~\cite{esser2024scalingrectifiedflowtransformers} & 0.42  & 0.44 &0.48 &0.39  &0.47 &0.29 & 0.42 \\
        & SDXL~\cite{2023SDXL} &0.43  & 0.48 &0.47  &0.44  &0.45 &0.27 & 0.43 \\
        
        &SD3.5-Large~\cite{esser2024scalingrectifiedflowtransformers} & 0.44 &0.50 &0.58  & 0.44&0.52 &0.31 & 0.46 \\

        & PixArt-$\alpha$~\cite{chen2023pixart} &  0.45  & 0.50& 0.48 & 0.49&0.56 &0.34 &  0.47\\

        & FLUX.1-dev~\cite{flux2024} &0.48  & 0.58 &0.62 &0.42  &0.51 & 0.35& 0.50 \\
        
        \midrule
        \multirow{6}{*}{\textit{Unified}}
& Janus~\cite{wu2024janus} &0.16 &0.26 &0.35 & 0.28 &0.30 & 0.14&  0.23\\
& Janus-Pro-1.5B~\cite{chen2025janus} & 0.20& 0.28&0.45 & 0.24 & 0.32& 0.16&  0.26\\
& Orthus~\cite{kou2024orthus} &0.23 &0.31 &0.38 &0.28  & 0.31&0.20 &  0.27\\

&VILA-U~\cite{wu2024vila}& 0.26 &0.33  & 0.37 &0.35  &0.39 &0.23 &  0.31\\
& Show-o~\cite{xie2024show} & \underline{0.28} &\underline{0.40}  &\underline{0.48} & \underline{0.30}&\textbf{0.46} &\textbf{0.30} &  \underline{0.35}\\

& \textbf{Harmon-1.5B} &  \textbf{0.38} & \textbf{0.48} & \textbf{0.52} & \textbf{0.37} & \underline{0.44} & \underline{0.29} & \textbf{0.41} \\
        \bottomrule
    \end{tabular}

    }
\label{tab:wise}
\end{table*}

\begin{figure*}[t]
  \centering
\includegraphics[width=1.0\textwidth]{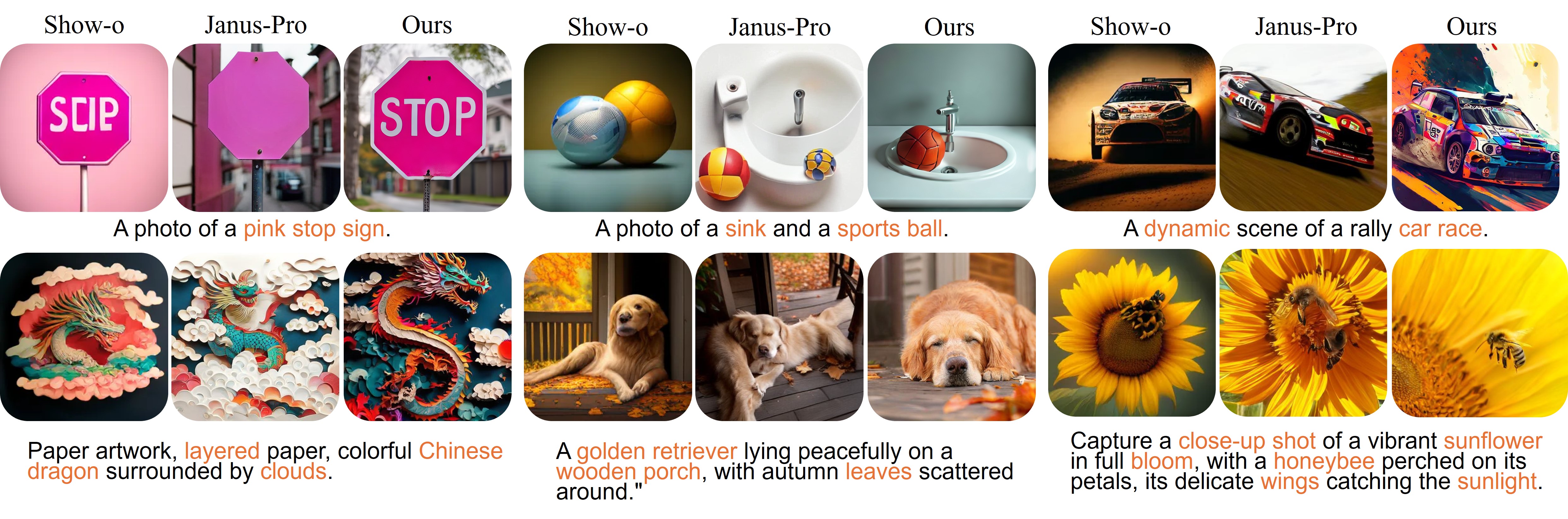}
  \vspace{-20pt}
  \caption{Qualitative comparison between Show-o (1.3B), Janus-Pro (1.5B) and our Harmon-1.5B on text-to-image generation. The text below each image represents the generation prompt, with key terms guiding the generation highlighted in \textcolor{orange}{orange}. Best viewed on screen.}
  \label{fig:vis_compare}
\end{figure*}

\noindent\textbf{Qualitative Comparison.} In Figure~\ref{fig:vis_compare}, we visually compare our generation results with state-of-the-art unified models, \ie, Show-o~\cite{xie2024show} and Janus-Pro~\cite{chen2025janus}. For a fair comparison, we select Show-o-1.3B-512, Janus-Pro-1.5B and our Harmon-1.5B. We observe better consistency between images and prompts in the results of Harmon. For example, Harmon is able to precisely generate the `pink stop sign' while Show-o and Janus-Pro fail to generate the textual contents. Harmon also successfully handles the `sink' that is missed in Show-o's generation result. Moreover, Harmon tends to produce images of higher visual quality, such as the `paper artwork' and `dog' in the bottom left and bottom middle examples, where Show-o and Janus-Pro generate distorted results. We provide more generation results and comparisons with more models in the appendix.

\subsection{Analysis \& Ablation}

In this section, we study the effectiveness of our design choices. All experiments are based on Harmon-0.5B.

\noindent\textbf{Encoders for Visual Understanding.} We examine the potential of visual encoders for understanding tasks, including VQGAN, VAE and MAR, which are trained by generative losses. Additonally, we include a semantic encoder, SigLIP~\cite{zhai2023sigmoid}, which is trained by contrastive vision-language aligning. For a fair comparison, MAR-B and SigLIP-Base-Patch16 are chosen. As shown in Table~\ref{tab:ablation_encoder}, VQGAN and VAE, which focus on image intrinsic features like local details and textures for high-fidelity compression, lag behind MAR and SigLIP by large margins. Meanwhile, MAR encoder performs comparably with SigLip, a language-aligned semantic encoder. The results are consistent with findings in MAE~\cite{he2022masked} that visual encoders can learn essential visual semantics and concepts through mask-ratio MiM training. In addition, the benchmark results also conform with the linear probing accuracies observed in our preliminary study (Figure~\ref{fig:pilot_study}).

\begin{table*}[t]
\centering

\begin{minipage}[t]{1.0\textwidth}
  \centering
\caption{Choices of visual encoders for understanding. `Acc' represents the top-5 linear probing accuracy on ImageNet~\cite{deng2009imagenet}.}
\vspace{-4pt}
\scalebox{0.87}
 {

\begin{tabular}{l|c|c|ccccccc}
\hline
\textbf{\#} &\textbf{Encoder}  &\textbf{Acc}$\uparrow$ 
 &\textbf{POPE}$\uparrow$ 
 &\textbf{MME-P}$\uparrow$ 
  &\textbf{MME-C}$\uparrow$ 
   &\textbf{MMB}$\uparrow$
    &\textbf{SEED}$\uparrow$ 
 &\textbf{GQA}$\uparrow$
 &\textbf{MMMU}$\uparrow$

\\ 
\hline
\rowcolor{gray!20} 1 & SigLIP & 95.9 &85.1 & 1203 & 258& 61.1 & 63.2 & 56.1 & 35.2   \\
2 & VQGAN  & 18.2 & 57.2 & 67.3  &  21.8 & 37.3 & 38.3& 38.0& 27.7\\
3 & VAE &18.5 & 63.8 & 732 & 223 &44.9& 42.5& 40.2 &  30.3   \\
4 & MAR  & \textbf{83.1} & \textbf{86.1} & \textbf{1123} & \textbf{262} & \textbf{60.1} & \textbf{62.2} & \textbf{55.7} & \textbf{33.3} \\ 


\bottomrule
\end{tabular}}
\label{tab:ablation_encoder}
\end{minipage}

\begin{minipage}[t]{1.0\textwidth}
  \centering
  \vspace{4pt}
\caption{Ablation study on the effects of the three training stages.}
\vspace{-4pt}
\scalebox{0.85}
 {\begin{tabular}{l|ccc|ccccc|ccc}
\hline
 \multirow{2}{*}{\textbf{\#}} &\multicolumn{3}{c|}{\textbf{Stage}} 
 &\multirow{2}{*}{\textbf{POPE}$\uparrow$} 
 &\multirow{2}{*}{\textbf{MME-P}$\uparrow$} 
  &\multirow{2}{*}{\textbf{MME-C}$\uparrow$} 
 &\multirow{2}{*}{\textbf{GQA}$\uparrow$} &\multirow{2}{*}{\textbf{MMMU}$\uparrow$} 
 &\multirow{2}{*}{\textbf{MSCOCO}$\downarrow$}
 &\multirow{2}{*}{\textbf{MJHQ}$\downarrow$}
  &\multirow{2}{*}{\textbf{GenEval}$\uparrow$}\\ 
 & \textbf{1} &  \textbf{2} & \textbf{3}&&&&&&
\\ 
\hline
1 & \xmark & \cmark  & \cmark& 85.2 & 1003 & 218  & 51.4  & 33.2 & 10.23 & 7.56 & 0.66 \\ 
2 & \cmark & \xmark & \cmark& 84.7 & 1064 &217 &50.2 & 32.8 & 10.99  & 8.12 & 0.65 \\
3 & \cmark & \cmark & \xmark& 85.6 & 1111 & 251& 54.1& 34.0 & 15.64 & 16.85 &0.56   \\
4 & \cmark & \cmark & \cmark& \textbf{86.5} & \textbf{1148} & \textbf{260} & \textbf{56.3} & \textbf{34.2}& \textbf{8.86}&\textbf{6.08}& \textbf{0.71}  \\ 

\bottomrule
\end{tabular}}
\label{tab:ablation_stages}
\end{minipage}

\begin{minipage}[t]{1.0\textwidth}
  \centering
  \vspace{4pt}
\caption{Ablation study on the effect of image resolution in stage III.}
\vspace{-4pt}
\scalebox{0.85}
 {\begin{tabular}{l|c|ccccc|ccc}
\hline
\textbf{\#} &\textbf{Resolution}& \textbf{POPE}$\uparrow$ & \textbf{MME-P}$\uparrow$ & \textbf{MME-C}$\uparrow$ & \textbf{GQA}$\uparrow$ & \textbf{MMMU}$\uparrow$ & \textbf{MSCOCO}$\downarrow$&\textbf{MJHQ}$\downarrow$& \textbf{GenEval}$\uparrow$ \\ 
\hline
1 & 256  & 86.1 & 1120 &258 & 55.4 & 32.6  & 11.50 & 9.91 &0.68 \\ 
2 & 384 & 86.5& 1144 & 260 & 55.5& 33.7&  10.97 &8.65 & 0.69  \\ 
3 & 512 & \textbf{86.5} & \textbf{1148} & \textbf{260} & \textbf{56.3} & \textbf{34.2}& \textbf{8.86}&\textbf{6.08}& \textbf{0.71}    \\

\bottomrule
\end{tabular}}
\label{tab:ablation_image_res}
\end{minipage}

\end{table*}

\noindent\textbf{Training Stages.} In Table~\ref{tab:ablation_stages}, we study the impact of each training stage. By comparing rows \#1, \#2 \& \#3 with row \#4, we observe that all training stages contribute to the overall performance. Specifically,  the alignment in stage I (\#1 and \#4) and comprehensive training in stage II (\#2 and \#4) mainly affect the performance on understanding benchmarks while the high-quality training in stage III, which features aesthetical images with higher resolutions (\#3 and \#4), improves generation performances the most, especially on the MJHQ-30K benchmark that measures visual quality.

\noindent\textbf{Image Resolutions.} In Table~\ref{tab:ablation_image_res}, we study the effect of increasing image resolution in the final training stage. Overall, performances on both understanding and generation tasks increase with higher resolutions. Specifically, the MSCOCO-30K and MJHQ-30K benchmarks that demand visual quality are more sensitive to image resolutions, where FID values drop drastically when image resolution increases. In contrast, performance gains are less pronounced on the general image understanding tasks as well as the GenEval benchmark that weighs more on alignment between images and texts in instructed-based generations.

\begin{figure}[t]
  \centering
\includegraphics[width=0.48\textwidth]{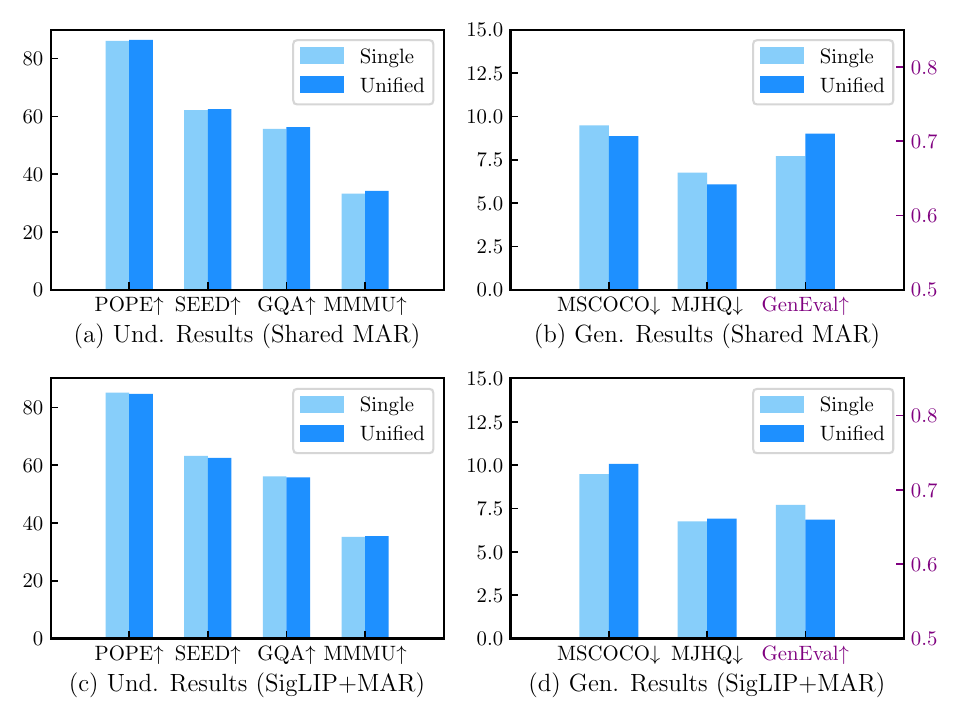}
  \vspace{-12pt}
  \caption{Ablation study on the mutual effect between understanding and generation supervision. Our shared encoding design brings performance gains on image generation (b) while maintaining understanding capability (a).} \label{fig:ablation_single_vs_unified}
  \vspace{-10pt}
\end{figure}

\noindent\textbf{Single Task v.s. Unified Training.}
 We study the mutual effect between generation and understanding by training with single-task losses. As shown in Figure~\ref{fig:ablation_single_vs_unified}(a), the unified training of the two tasks does not decrease scores on understanding tasks compared with single-task training. For image generation, we observe that the understanding loss even boosts generation performance, especially on the GenEval benchmark that measures the consistency between generated images and user instructions, as suggested by Figure~\ref{fig:ablation_single_vs_unified}(b). In addition, we also compare our shared encoding strategy with a Janus-like design, \ie, using a semantic encoder (SigLIP-Base-Patch16~\cite{zhai2023sigmoid}) for understanding and MAR encoder for generation. As shown in Figure~\ref{fig:ablation_single_vs_unified} (c,d), the decoupled encoding architecture accommodates the two tasks well. However, the synergistic effect disappears, suggesting that a shared representation is key to the co-evolution of understanding and generation capabilities within a unified model. 

\section{Conclusion}
In this paper, we investigate the visual encoders in unified understanding and generation frameworks. Through a preliminary study including linear probing and feature visualization, we recognize the potential of MAR for visual understanding beyond its in-born generative role. Based on this finding, we present Harmon, a unified framework that harmonizes image understanding and generation in MAR's encoding space. Through a progressive three-stage training procedure, Harmon obtains state-of-the-art results on text-to-image generation benchmarks while achieving competitive performance on multimodal question-answering benchmarks. Moreover, we observe the synergy between the two tasks in our experiment, where the co-training with understanding loss boosts generation performance.

\paragraph{Acknowledgement.} This research is supported by the National Research Foundation, Singapore under its AI Singapore Programme (AISG Award No: AISG3-PhD-2023-08-048T), the RIE2020 Industry Alignment Fund – Industry Collaboration Projects (IAF-ICP) Funding Initiative, as well as cash and in-kind contribution from the industry partner(s).


{
    \small
    \bibliographystyle{ieeenat_fullname}
    \bibliography{main}

\begin{thebibliography}{100}
\providecommand{\natexlab}[1]{#1}
\providecommand{\url}[1]{\texttt{#1}}
\expandafter\ifx\csname urlstyle\endcsname\relax
  \providecommand{\doi}[1]{doi: #1}\else
  \providecommand{\doi}{doi: \begingroup \urlstyle{rm}\Url}\fi

\bibitem[Bai et~al.(2024)Bai, Ye, Chow, Song, Chen, Li, Dong, Zhu, and Shuicheng]{bai2024meissonic}
Jinbin Bai, Tian Ye, Wei Chow, Enxin Song, Qing-Guo Chen, Xiangtai Li, Zhen Dong, Lei Zhu, and YAN Shuicheng.
\newblock Meissonic: Revitalizing masked generative transformers for efficient high-resolution text-to-image synthesis.
\newblock In \emph{The Thirteenth International Conference on Learning Representations}, 2024.

\bibitem[Bao et~al.(2021)Bao, Dong, Piao, and Wei]{bao2021beit}
Hangbo Bao, Li Dong, Songhao Piao, and Furu Wei.
\newblock Beit: Bert pre-training of image transformers.
\newblock \emph{arXiv preprint arXiv:2106.08254}, 2021.

\bibitem[Betker et~al.(2023)Betker, Goh, Jing, Brooks, Wang, Li, Ouyang, Zhuang, Lee, Guo, et~al.]{dalle3}
James Betker, Gabriel Goh, Li Jing, Tim Brooks, Jianfeng Wang, Linjie Li, Long Ouyang, Juntang Zhuang, Joyce Lee, Yufei Guo, et~al.
\newblock Improving image generation with better captions.
\newblock \emph{Computer Science. https://cdn. openai. com/papers/dall-e-3. pdf}, 2\penalty0 (3):\penalty0 8, 2023.

\bibitem[Brown(2020)]{brown2020language}
Tom~B Brown.
\newblock Language models are few-shot learners.
\newblock \emph{arXiv preprint arXiv:2005.14165}, 2020.

\bibitem[Chang et~al.(2022)Chang, Zhang, Jiang, Liu, and Freeman]{chang2022maskgit}
Huiwen Chang, Han Zhang, Lu Jiang, Ce Liu, and William~T Freeman.
\newblock Maskgit: Masked generative image transformer.
\newblock In \emph{Proceedings of the IEEE/CVF Conference on Computer Vision and Pattern Recognition}, pages 11315--11325, 2022.

\bibitem[Chang et~al.(2023)Chang, Zhang, Barber, Maschinot, Lezama, Jiang, Yang, Murphy, Freeman, Rubinstein, et~al.]{chang2023muse}
Huiwen Chang, Han Zhang, Jarred Barber, AJ Maschinot, Jose Lezama, Lu Jiang, Ming-Hsuan Yang, Kevin Murphy, William~T Freeman, Michael Rubinstein, et~al.
\newblock Muse: Text-to-image generation via masked generative transformers.
\newblock \emph{arXiv preprint arXiv:2301.00704}, 2023.

\bibitem[Changpinyo et~al.(2021)Changpinyo, Sharma, Ding, and Soricut]{changpinyo2021conceptual}
Soravit Changpinyo, Piyush Sharma, Nan Ding, and Radu Soricut.
\newblock Conceptual 12m: Pushing web-scale image-text pre-training to recognize long-tail visual concepts.
\newblock In \emph{Proceedings of the IEEE/CVF conference on computer vision and pattern recognition}, pages 3558--3568, 2021.

\bibitem[Chattopadhay et~al.(2018)Chattopadhay, Sarkar, Howlader, and Balasubramanian]{chattopadhay2018grad}
Aditya Chattopadhay, Anirban Sarkar, Prantik Howlader, and Vineeth~N Balasubramanian.
\newblock Grad-cam++: Generalized gradient-based visual explanations for deep convolutional networks.
\newblock In \emph{2018 IEEE winter conference on applications of computer vision (WACV)}, pages 839--847. IEEE, 2018.

\bibitem[Chen et~al.(2023{\natexlab{a}})Chen, Yu, Ge, Yao, Xie, Wu, Wang, Kwok, Luo, Lu, et~al.]{2023Pixelartalpha}
Junsong Chen, Jincheng Yu, Chongjian Ge, Lewei Yao, Enze Xie, Yue Wu, Zhongdao Wang, James Kwok, Ping Luo, Huchuan Lu, et~al.
\newblock {PixArt-alpha}: Fast training of diffusion transformer for photorealistic text-to-image synthesis.
\newblock \emph{arXiv preprint arXiv:2310.00426}, 2023{\natexlab{a}}.

\bibitem[Chen et~al.(2023{\natexlab{b}})Chen, Yu, Ge, Yao, Xie, Wu, Wang, Kwok, Luo, Lu, et~al.]{chen2023pixart}
Junsong Chen, Jincheng Yu, Chongjian Ge, Lewei Yao, Enze Xie, Yue Wu, Zhongdao Wang, James Kwok, Ping Luo, Huchuan Lu, et~al.
\newblock Pixart-$alpha$: Fast training of diffusion transformer for photorealistic text-to-image synthesis.
\newblock \emph{arXiv preprint arXiv:2310.00426}, 2023{\natexlab{b}}.

\bibitem[Chen et~al.(2024{\natexlab{a}})Chen, Ge, Xie, Wu, Yao, Ren, Wang, Luo, Lu, and Li]{2024pixartsigma}
Junsong Chen, Chongjian Ge, Enze Xie, Yue Wu, Lewei Yao, Xiaozhe Ren, Zhongdao Wang, Ping Luo, Huchuan Lu, and Zhenguo Li.
\newblock {PixArt-Sigma}: Weak-to-strong training of diffusion transformer for {4K} text-to-image generation.
\newblock \emph{arXiv preprint arXiv:2403.04692}, 2024{\natexlab{a}}.

\bibitem[Chen et~al.(2023{\natexlab{c}})Chen, Li, Dong, Zhang, He, Wang, Zhao, and Lin]{chen2023sharegpt4v}
Lin Chen, Jisong Li, Xiaoyi Dong, Pan Zhang, Conghui He, Jiaqi Wang, Feng Zhao, and Dahua Lin.
\newblock Sharegpt4v: Improving large multi-modal models with better captions.
\newblock \emph{arXiv preprint arXiv:2311.12793}, 2023{\natexlab{c}}.

\bibitem[Chen et~al.(2024{\natexlab{b}})Chen, Li, Dong, Zhang, He, Wang, Zhao, and Lin]{chen2024sharegpt4v}
Lin Chen, Jinsong Li, Xiaoyi Dong, Pan Zhang, Conghui He, Jiaqi Wang, Feng Zhao, and Dahua Lin.
\newblock Sharegpt4v: Improving large multi-modal models with better captions.
\newblock In \emph{European Conference on Computer Vision}, pages 370--387. Springer, 2024{\natexlab{b}}.

\bibitem[Chen et~al.(2015)Chen, Fang, Lin, Vedantam, Gupta, Doll{\'a}r, and Zitnick]{chen2015microsoft}
Xinlei Chen, Hao Fang, Tsung-Yi Lin, Ramakrishna Vedantam, Saurabh Gupta, Piotr Doll{\'a}r, and C~Lawrence Zitnick.
\newblock Microsoft coco captions: Data collection and evaluation server.
\newblock \emph{arXiv preprint arXiv:1504.00325}, 2015.

\bibitem[Chen et~al.(2025)Chen, Wu, Liu, Pan, Liu, Xie, Yu, and Ruan]{chen2025janus}
Xiaokang Chen, Zhiyu Wu, Xingchao Liu, Zizheng Pan, Wen Liu, Zhenda Xie, Xingkai Yu, and Chong Ruan.
\newblock Janus-pro: Unified multimodal understanding and generation with data and model scaling.
\newblock \emph{arXiv preprint arXiv:2501.17811}, 2025.

\bibitem[Chen et~al.(2024{\natexlab{c}})Chen, Wang, Tian, Ye, Gao, Cui, Tong, Hu, Luo, Ma, et~al.]{chen2024far}
Zhe Chen, Weiyun Wang, Hao Tian, Shenglong Ye, Zhangwei Gao, Erfei Cui, Wenwen Tong, Kongzhi Hu, Jiapeng Luo, Zheng Ma, et~al.
\newblock How far are we to gpt-4v? closing the gap to commercial multimodal models with open-source suites.
\newblock \emph{arXiv preprint arXiv:2404.16821}, 2024{\natexlab{c}}.

\bibitem[Chen et~al.(2024{\natexlab{d}})Chen, Wu, Wang, Su, Chen, Xing, Zhong, Zhang, Zhu, Lu, et~al.]{chen2024internvl}
Zhe Chen, Jiannan Wu, Wenhai Wang, Weijie Su, Guo Chen, Sen Xing, Muyan Zhong, Qinglong Zhang, Xizhou Zhu, Lewei Lu, et~al.
\newblock Internvl: Scaling up vision foundation models and aligning for generic visual-linguistic tasks.
\newblock In \emph{Proceedings of the IEEE/CVF Conference on Computer Vision and Pattern Recognition}, pages 24185--24198, 2024{\natexlab{d}}.

\bibitem[Cherti et~al.(2023)Cherti, Beaumont, Wightman, Wortsman, Ilharco, Gordon, Schuhmann, Schmidt, and Jitsev]{cherti2023reproducible}
Mehdi Cherti, Romain Beaumont, Ross Wightman, Mitchell Wortsman, Gabriel Ilharco, Cade Gordon, Christoph Schuhmann, Ludwig Schmidt, and Jenia Jitsev.
\newblock Reproducible scaling laws for contrastive language-image learning.
\newblock In \emph{Proceedings of the IEEE/CVF conference on computer vision and pattern recognition}, pages 2818--2829, 2023.

\bibitem[Chu et~al.(2023)Chu, Qiao, Lin, Xu, Yang, Hu, Wei, Zhang, Zhang, Wei, et~al.]{chu2023mobilevlm}
Xiangxiang Chu, Limeng Qiao, Xinyang Lin, Shuang Xu, Yang Yang, Yiming Hu, Fei Wei, Xinyu Zhang, Bo Zhang, Xiaolin Wei, et~al.
\newblock Mobilevlm: A fast, reproducible and strong vision language assistant for mobile devices.
\newblock \emph{arXiv preprint arXiv:2312.16886}, 2023.

\bibitem[Chu et~al.(2024)Chu, Qiao, Zhang, Xu, Wei, Yang, Sun, Hu, Lin, Zhang, et~al.]{chu2024mobilevlm2}
Xiangxiang Chu, Limeng Qiao, Xinyu Zhang, Shuang Xu, Fei Wei, Yang Yang, Xiaofei Sun, Yiming Hu, Xinyang Lin, Bo Zhang, et~al.
\newblock Mobilevlm v2: Faster and stronger baseline for vision language model.
\newblock \emph{arXiv preprint arXiv:2402.03766}, 2024.

\bibitem[Dai et~al.(2023)Dai, Li, Li, Tiong, Zhao, Wang, Li, Fung, and Hoi]{instructblip}
Wenliang Dai, Junnan Li, Dongxu Li, Anthony Meng~Huat Tiong, Junqi Zhao, Weisheng Wang, Boyang Li, Pascale Fung, and Steven Hoi.
\newblock Instructblip: Towards general-purpose vision-language models with instruction tuning, 2023.

\bibitem[dclure(2022)]{dclure_laion_aesthetics_12m_umap}
dclure.
\newblock Laion-aesthetics-umap.
\newblock \url{https://huggingface.co/datasets/dclure/laion-aesthetics-12m-umap}, 2022.

\bibitem[Deng et~al.(2009)Deng, Dong, Socher, Li, Li, and Fei-Fei]{deng2009imagenet}
Jia Deng, Wei Dong, Richard Socher, Li-Jia Li, Kai Li, and Li Fei-Fei.
\newblock Imagenet: A large-scale hierarchical image database.
\newblock In \emph{2009 IEEE conference on computer vision and pattern recognition}, pages 248--255. Ieee, 2009.

\bibitem[Desai et~al.(2021)Desai, Kaul, Aysola, and Johnson]{desai2021redcaps}
Karan Desai, Gaurav Kaul, Zubin Aysola, and Justin Johnson.
\newblock Redcaps: Web-curated image-text data created by the people, for the people.
\newblock \emph{arXiv preprint arXiv:2111.11431}, 2021.

\bibitem[Emporium(2024)]{flickr-megalith-10m-internvl2-multi-caption}
Caption Emporium.
\newblock flickr-megalith-10m-internvl2-multi-caption.
\newblock \url{https://huggingface.co/datasets/CaptionEmporium/flickr-megalith-10m-internvl2-multi-caption}, 2024.

\bibitem[Esser et~al.(2021)Esser, Rombach, and Ommer]{esser2021taming}
Patrick Esser, Robin Rombach, and Bjorn Ommer.
\newblock Taming transformers for high-resolution image synthesis.
\newblock In \emph{Proceedings of the IEEE/CVF conference on computer vision and pattern recognition}, pages 12873--12883, 2021.

\bibitem[Esser et~al.(2024)Esser, Kulal, Blattmann, Entezari, Müller, Saini, Levi, Lorenz, Sauer, Boesel, Podell, Dockhorn, English, Lacey, Goodwin, Marek, and Rombach]{esser2024scalingrectifiedflowtransformers}
Patrick Esser, Sumith Kulal, Andreas Blattmann, Rahim Entezari, Jonas Müller, Harry Saini, Yam Levi, Dominik Lorenz, Axel Sauer, Frederic Boesel, Dustin Podell, Tim Dockhorn, Zion English, Kyle Lacey, Alex Goodwin, Yannik Marek, and Robin Rombach.
\newblock Scaling rectified flow transformers for high-resolution image synthesis, 2024.

\bibitem[Fan et~al.(2024)Fan, Li, Qin, Li, Sun, Rubinstein, Sun, He, and Tian]{fan2024fluid}
Lijie Fan, Tianhong Li, Siyang Qin, Yuanzhen Li, Chen Sun, Michael Rubinstein, Deqing Sun, Kaiming He, and Yonglong Tian.
\newblock Fluid: Scaling autoregressive text-to-image generative models with continuous tokens.
\newblock \emph{arXiv preprint arXiv:2410.13863}, 2024.

\bibitem[Fu et~al.(2023)Fu, Chen, Shen, Qin, Zhang, Lin, Yang, Zheng, Li, Sun, et~al.]{fu2023mme}
Chaoyou Fu, Peixian Chen, Yunhang Shen, Yulei Qin, Mengdan Zhang, Xu Lin, Jinrui Yang, Xiawu Zheng, Ke Li, Xing Sun, et~al.
\newblock Mme: A comprehensive evaluation benchmark for multimodal large language models.
\newblock \emph{arXiv preprint arXiv:2306.13394}, 2023.

\bibitem[Gadre et~al.(2023)Gadre, Ilharco, Fang, Hayase, Smyrnis, Nguyen, Marten, Wortsman, Ghosh, Zhang, et~al.]{gadre2023datacomp}
Samir~Yitzhak Gadre, Gabriel Ilharco, Alex Fang, Jonathan Hayase, Georgios Smyrnis, Thao Nguyen, Ryan Marten, Mitchell Wortsman, Dhruba Ghosh, Jieyu Zhang, et~al.
\newblock Datacomp: In search of the next generation of multimodal datasets.
\newblock \emph{Advances in Neural Information Processing Systems}, 36:\penalty0 27092--27112, 2023.

\bibitem[Ge et~al.(2024)Ge, Zhao, Zhu, Ge, Yi, Song, Li, Ding, and Shan]{ge2024seed}
Yuying Ge, Sijie Zhao, Jinguo Zhu, Yixiao Ge, Kun Yi, Lin Song, Chen Li, Xiaohan Ding, and Ying Shan.
\newblock Seed-x: Multimodal models with unified multi-granularity comprehension and generation.
\newblock \emph{arXiv preprint arXiv:2404.14396}, 2024.

\bibitem[Ghosh et~al.(2024)Ghosh, Hajishirzi, and Schmidt]{ghosh2024geneval}
Dhruba Ghosh, Hannaneh Hajishirzi, and Ludwig Schmidt.
\newblock Geneval: An object-focused framework for evaluating text-to-image alignment.
\newblock \emph{Advances in Neural Information Processing Systems}, 36, 2024.

\bibitem[Gu et~al.(2024)Gu, Zhang, Zhou, Yu, Xing, Wang, Cao, Jia, Zhang, Wang, et~al.]{gu2024infinity}
Shuhao Gu, Jialing Zhang, Siyuan Zhou, Kevin Yu, Zhaohu Xing, Liangdong Wang, Zhou Cao, Jintao Jia, Zhuoyi Zhang, Yixuan Wang, et~al.
\newblock Infinity-mm: Scaling multimodal performance with large-scale and high-quality instruction data.
\newblock \emph{arXiv preprint arXiv:2410.18558}, 2024.

\bibitem[He et~al.(2022)He, Chen, Xie, Li, Doll{\'a}r, and Girshick]{he2022masked}
Kaiming He, Xinlei Chen, Saining Xie, Yanghao Li, Piotr Doll{\'a}r, and Ross Girshick.
\newblock Masked autoencoders are scalable vision learners.
\newblock In \emph{Proceedings of the IEEE/CVF conference on computer vision and pattern recognition}, pages 16000--16009, 2022.

\bibitem[Ho and Salimans(2022)]{ho2022classifier}
Jonathan Ho and Tim Salimans.
\newblock Classifier-free diffusion guidance.
\newblock \emph{arXiv preprint arXiv:2207.12598}, 2022.

\bibitem[Ho et~al.(2020)Ho, Jain, and Abbeel]{ho2020denoising}
Jonathan Ho, Ajay Jain, and Pieter Abbeel.
\newblock Denoising diffusion probabilistic models.
\newblock \emph{Advances in neural information processing systems}, 33:\penalty0 6840--6851, 2020.

\bibitem[Hudson and Manning(2019)]{hudson2019gqa}
Drew~A Hudson and Christopher~D Manning.
\newblock Gqa: A new dataset for real-world visual reasoning and compositional question answering.
\newblock In \emph{Proceedings of the IEEE/CVF conference on computer vision and pattern recognition}, pages 6700--6709, 2019.

\bibitem[jackyhate(2024)]{t2i2m}
jackyhate.
\newblock text-to-image-2m.
\newblock \url{https://huggingface.co/datasets/jackyhate/text-to-image-2M}, 2024.

\bibitem[Kang et~al.(2023)Kang, Zhu, Zhang, Park, Shechtman, Paris, and Park]{kang2023scaling}
Minguk Kang, Jun-Yan Zhu, Richard Zhang, Jaesik Park, Eli Shechtman, Sylvain Paris, and Taesung Park.
\newblock Scaling up gans for text-to-image synthesis.
\newblock In \emph{Proceedings of the IEEE/CVF Conference on Computer Vision and Pattern Recognition}, pages 10124--10134, 2023.

\bibitem[Kingma et~al.(2013)Kingma, Welling, et~al.]{kingma2013auto}
Diederik~P Kingma, Max Welling, et~al.
\newblock Auto-encoding variational bayes, 2013.

\bibitem[Kou et~al.(2024)Kou, Jin, Liu, Ma, Jia, Chen, Jiang, and Deng]{kou2024orthus}
Siqi Kou, Jiachun Jin, Chang Liu, Ye Ma, Jian Jia, Quan Chen, Peng Jiang, and Zhijie Deng.
\newblock Orthus: Autoregressive interleaved image-text generation with modality-specific heads.
\newblock \emph{arXiv preprint arXiv:2412.00127}, 2024.

\bibitem[Labs(2024)]{flux2024}
Black~Forest Labs.
\newblock Flux.
\newblock \url{https://github.com/black-forest-labs/flux}, 2024.

\bibitem[Li et~al.(2023{\natexlab{a}})Li, Wang, Wang, Ge, Ge, and Shan]{li2023seed}
Bohao Li, Rui Wang, Guangzhi Wang, Yuying Ge, Yixiao Ge, and Ying Shan.
\newblock Seed-bench: Benchmarking multimodal llms with generative comprehension.
\newblock \emph{arXiv preprint arXiv:2307.16125}, 2023{\natexlab{a}}.

\bibitem[Li et~al.(2024{\natexlab{a}})Li, Zhang, Guo, Zhang, Li, Zhang, Zhang, Li, Liu, and Li]{li2024llava}
Bo Li, Yuanhan Zhang, Dong Guo, Renrui Zhang, Feng Li, Hao Zhang, Kaichen Zhang, Yanwei Li, Ziwei Liu, and Chunyuan Li.
\newblock Llava-onevision: Easy visual task transfer.
\newblock \emph{arXiv preprint arXiv:2408.03326}, 2024{\natexlab{a}}.

\bibitem[Li et~al.(2024{\natexlab{b}})Li, Kamko, Akhgari, Sabet, Xu, and Doshi]{li2024playground}
Daiqing Li, Aleks Kamko, Ehsan Akhgari, Ali Sabet, Linmiao Xu, and Suhail Doshi.
\newblock Playground v2. 5: Three insights towards enhancing aesthetic quality in text-to-image generation.
\newblock \emph{arXiv preprint arXiv:2402.17245}, 2024{\natexlab{b}}.

\bibitem[Li et~al.(2024{\natexlab{c}})Li, Tian, Li, Deng, and He]{li2024autoregressive}
Tianhong Li, Yonglong Tian, He Li, Mingyang Deng, and Kaiming He.
\newblock Autoregressive image generation without vector quantization.
\newblock \emph{arXiv preprint arXiv:2406.11838}, 2024{\natexlab{c}}.

\bibitem[Li et~al.(2024{\natexlab{d}})Li, Zhang, Diao, Wang, Wang, and Duan]{li2024densefusion}
Xiaotong Li, Fan Zhang, Haiwen Diao, Yueze Wang, Xinlong Wang, and Ling-Yu Duan.
\newblock Densefusion-1m: Merging vision experts for comprehensive multimodal perception.
\newblock \emph{arXiv preprint arXiv:2407.08303}, 2024{\natexlab{d}}.

\bibitem[Li et~al.(2023{\natexlab{b}})Li, Du, Zhou, Wang, Zhao, and Wen]{li2023evaluating}
Yifan Li, Yifan Du, Kun Zhou, Jinpeng Wang, Wayne~Xin Zhao, and Ji-Rong Wen.
\newblock Evaluating object hallucination in large vision-language models.
\newblock \emph{arXiv preprint arXiv:2305.10355}, 2023{\natexlab{b}}.

\bibitem[Li et~al.(2024{\natexlab{e}})Li, Zhang, Wang, Zhong, Chen, Chu, Liu, and Jia]{li2024mini}
Yanwei Li, Yuechen Zhang, Chengyao Wang, Zhisheng Zhong, Yixin Chen, Ruihang Chu, Shaoteng Liu, and Jiaya Jia.
\newblock Mini-gemini: Mining the potential of multi-modality vision language models.
\newblock \emph{arXiv preprint arXiv:2403.18814}, 2024{\natexlab{e}}.

\bibitem[Li et~al.(2024{\natexlab{f}})Li, Li, Shi, Farimani, Kluger, Yang, and Wang]{li2024dual}
Zijie Li, Henry Li, Yichun Shi, Amir~Barati Farimani, Yuval Kluger, Linjie Yang, and Peng Wang.
\newblock Dual diffusion for unified image generation and understanding.
\newblock \emph{arXiv preprint arXiv:2501.00289}, 2024{\natexlab{f}}.

\bibitem[Li et~al.(2024{\natexlab{g}})Li, Zhang, Lin, Xiong, Long, Deng, Zhang, Liu, Huang, Xiao, et~al.]{2024hunyuandit}
Zhimin Li, Jianwei Zhang, Qin Lin, Jiangfeng Xiong, Yanxin Long, Xinchi Deng, Yingfang Zhang, Xingchao Liu, Minbin Huang, Zedong Xiao, et~al.
\newblock {Hunyuan-DiT}: A powerful multi-resolution diffusion transformer with fine-grained chinese understanding.
\newblock \emph{arXiv preprint arXiv:2405.08748}, 2024{\natexlab{g}}.

\bibitem[Liu et~al.(2024{\natexlab{a}})Liu, Li, Li, and Lee]{liu2024improved}
Haotian Liu, Chunyuan Li, Yuheng Li, and Yong~Jae Lee.
\newblock Improved baselines with visual instruction tuning.
\newblock In \emph{Proceedings of the IEEE/CVF Conference on Computer Vision and Pattern Recognition}, pages 26296--26306, 2024{\natexlab{a}}.

\bibitem[Liu et~al.(2024{\natexlab{b}})Liu, Li, Wu, and Lee]{liu2024visual}
Haotian Liu, Chunyuan Li, Qingyang Wu, and Yong~Jae Lee.
\newblock Visual instruction tuning.
\newblock \emph{Advances in neural information processing systems}, 36, 2024{\natexlab{b}}.

\bibitem[Liu et~al.(2024{\natexlab{c}})Liu, Yan, Zaharia, and Abbeel]{liu2024world}
Hao Liu, Wilson Yan, Matei Zaharia, and Pieter Abbeel.
\newblock World model on million-length video and language with ringattention.
\newblock \emph{arXiv preprint arXiv:2402.08268}, 2024{\natexlab{c}}.

\bibitem[Liu et~al.(2023)Liu, Duan, Zhang, Li, Zhang, Zhao, Yuan, Wang, He, Liu, et~al.]{liu2023mmbench}
Yuan Liu, Haodong Duan, Yuanhan Zhang, Bo Li, Songyang Zhang, Wangbo Zhao, Yike Yuan, Jiaqi Wang, Conghui He, Ziwei Liu, et~al.
\newblock Mmbench: Is your multi-modal model an all-around player?
\newblock \emph{arXiv preprint arXiv:2307.06281}, 2023.

\bibitem[Liu et~al.(2022)Liu, Mao, Wu, Feichtenhofer, Darrell, and Xie]{liu2022convnet}
Zhuang Liu, Hanzi Mao, Chao-Yuan Wu, Christoph Feichtenhofer, Trevor Darrell, and Saining Xie.
\newblock A convnet for the 2020s.
\newblock In \emph{Proceedings of the IEEE/CVF conference on computer vision and pattern recognition}, pages 11976--11986, 2022.

\bibitem[Lu et~al.(2024)Lu, Liu, Zhang, Wang, Dong, Liu, Sun, Ren, Li, Sun, et~al.]{lu2024deepseek}
Haoyu Lu, Wen Liu, Bo Zhang, Bingxuan Wang, Kai Dong, Bo Liu, Jingxiang Sun, Tongzheng Ren, Zhuoshu Li, Yaofeng Sun, et~al.
\newblock Deepseek-vl: towards real-world vision-language understanding.
\newblock \emph{arXiv preprint arXiv:2403.05525}, 2024.

\bibitem[Luo et~al.(2024)Luo, Shi, Ge, Yang, Wang, and Shan]{luo2024open}
Zhuoyan Luo, Fengyuan Shi, Yixiao Ge, Yujiu Yang, Limin Wang, and Ying Shan.
\newblock Open-magvit2: An open-source project toward democratizing auto-regressive visual generation.
\newblock \emph{arXiv preprint arXiv:2409.04410}, 2024.

\bibitem[madebyollin(2024)]{madebyollin_megalith_10m}
madebyollin.
\newblock Megalith-huggingface.
\newblock \url{https://huggingface.co/datasets/madebyollin/megalith-10m}, 2024.

\bibitem[Meyer et~al.(2024)Meyer, Padgett, Miller, and Exline]{meyer2024public}
Jordan Meyer, Nick Padgett, Cullen Miller, and Laura Exline.
\newblock Public domain 12m: A highly aesthetic image-text dataset with novel governance mechanisms.
\newblock \emph{arXiv preprint arXiv:2410.23144}, 2024.

\bibitem[Niu et~al.(2025)Niu, Ning, Zheng, Lin, Jin, Liao, Ning, Zhu, and Yuan]{niu2025wise}
Yuwei Niu, Munan Ning, Mengren Zheng, Bin Lin, Peng Jin, Jiaqi Liao, Kunpeng Ning, Bin Zhu, and Li Yuan.
\newblock Wise: A world knowledge-informed semantic evaluation for text-to-image generation.
\newblock \emph{arXiv preprint arXiv:2503.07265}, 2025.

\bibitem[Pan et~al.(2023)Pan, Sun, Ge, Li, Duan, Wu, Zhang, Zhou, Qin, Wang, Dai, Qiao, and Li]{pan2023journeydb}
Junting Pan, Keqiang Sun, Yuying Ge, Hao Li, Haodong Duan, Xiaoshi Wu, Renrui Zhang, Aojun Zhou, Zipeng Qin, Yi Wang, Jifeng Dai, Yu Qiao, and Hongsheng Li.
\newblock Journeydb: A benchmark for generative image understanding, 2023.

\bibitem[Peebles and Xie(2023)]{peebles2023scalable}
William Peebles and Saining Xie.
\newblock Scalable diffusion models with transformers.
\newblock In \emph{Proceedings of the IEEE/CVF international conference on computer vision}, pages 4195--4205, 2023.

\bibitem[Podell et~al.(2024)Podell, English, Lacey, Blattmann, Dockhorn, M{\"u}ller, Penna, and Rombach]{2023SDXL}
Dustin Podell, Zion English, Kyle Lacey, Andreas Blattmann, Tim Dockhorn, Jonas M{\"u}ller, Joe Penna, and Robin Rombach.
\newblock {SDXL}: Improving latent diffusion models for high-resolution image synthesis.
\newblock In \emph{ICLR}, 2024.

\bibitem[Qu et~al.(2024)Qu, Zhang, Liu, Wang, Jiang, Gao, Ye, Du, Yuan, and Wu]{qu2024tokenflow}
Liao Qu, Huichao Zhang, Yiheng Liu, Xu Wang, Yi Jiang, Yiming Gao, Hu Ye, Daniel~K Du, Zehuan Yuan, and Xinglong Wu.
\newblock Tokenflow: Unified image tokenizer for multimodal understanding and generation.
\newblock \emph{arXiv preprint arXiv:2412.03069}, 2024.

\bibitem[Radford(2018)]{radford2018improving}
Alec Radford.
\newblock Improving language understanding by generative pre-training.
\newblock 2018.

\bibitem[Radford et~al.(2019)Radford, Wu, Child, Luan, Amodei, Sutskever, et~al.]{radford2019language}
Alec Radford, Jeffrey Wu, Rewon Child, David Luan, Dario Amodei, Ilya Sutskever, et~al.
\newblock Language models are unsupervised multitask learners.
\newblock \emph{OpenAI blog}, 1\penalty0 (8):\penalty0 9, 2019.

\bibitem[Radford et~al.(2021)Radford, Kim, Hallacy, Ramesh, Goh, Agarwal, Sastry, Askell, Mishkin, Clark, et~al.]{radford2021learning}
Alec Radford, Jong~Wook Kim, Chris Hallacy, Aditya Ramesh, Gabriel Goh, Sandhini Agarwal, Girish Sastry, Amanda Askell, Pamela Mishkin, Jack Clark, et~al.
\newblock Learning transferable visual models from natural language supervision.
\newblock In \emph{International conference on machine learning}, pages 8748--8763. PMLR, 2021.

\bibitem[Ramesh et~al.(2021)Ramesh, Pavlov, Goh, Gray, Voss, Radford, Chen, and Sutskever]{ramesh2021zero}
Aditya Ramesh, Mikhail Pavlov, Gabriel Goh, Scott Gray, Chelsea Voss, Alec Radford, Mark Chen, and Ilya Sutskever.
\newblock Zero-shot text-to-image generation.
\newblock In \emph{International conference on machine learning}, pages 8821--8831. Pmlr, 2021.

\bibitem[Ramesh et~al.(2022)Ramesh, Dhariwal, Nichol, Chu, and Chen]{ramesh2022hierarchical}
Aditya Ramesh, Prafulla Dhariwal, Alex Nichol, Casey Chu, and Mark Chen.
\newblock Hierarchical text-conditional image generation with clip latents.
\newblock \emph{arXiv preprint arXiv:2204.06125}, 1\penalty0 (2):\penalty0 3, 2022.

\bibitem[Rombach et~al.(2022{\natexlab{a}})Rombach, Blattmann, Lorenz, Esser, and Ommer]{2022LDM}
Robin Rombach, Andreas Blattmann, Dominik Lorenz, Patrick Esser, and Bj{\"o}rn Ommer.
\newblock High-resolution image synthesis with latent diffusion models.
\newblock In \emph{CVPR}, 2022{\natexlab{a}}.

\bibitem[Rombach et~al.(2022{\natexlab{b}})Rombach, Blattmann, Lorenz, Esser, and Ommer]{rombach2022high}
Robin Rombach, Andreas Blattmann, Dominik Lorenz, Patrick Esser, and Bj{\"o}rn Ommer.
\newblock High-resolution image synthesis with latent diffusion models.
\newblock In \emph{Proceedings of the IEEE/CVF conference on computer vision and pattern recognition}, pages 10684--10695, 2022{\natexlab{b}}.

\bibitem[Schuhmann et~al.(2022)Schuhmann, Beaumont, Vencu, Gordon, Wightman, Cherti, Coombes, Katta, Mullis, Wortsman, et~al.]{schuhmann2022laion}
Christoph Schuhmann, Romain Beaumont, Richard Vencu, Cade Gordon, Ross Wightman, Mehdi Cherti, Theo Coombes, Aarush Katta, Clayton Mullis, Mitchell Wortsman, et~al.
\newblock Laion-5b: An open large-scale dataset for training next generation image-text models.
\newblock \emph{Advances in neural information processing systems}, 35:\penalty0 25278--25294, 2022.

\bibitem[Shi et~al.(2024)Shi, Han, Zhou, Liang, Lin, Zettlemoyer, and Yu]{shi2024llamafusion}
Weijia Shi, Xiaochuang Han, Chunting Zhou, Weixin Liang, Xi~Victoria Lin, Luke Zettlemoyer, and Lili Yu.
\newblock Llamafusion: Adapting pretrained language models for multimodal generation.
\newblock \emph{arXiv preprint arXiv:2412.15188}, 2024.

\bibitem[Singla et~al.(2024)Singla, Yue, Paul, Shirkavand, Jayawardhana, Ganjdanesh, Huang, Bhatele, Somepalli, and Goldstein]{singla2024pixels}
Vasu Singla, Kaiyu Yue, Sukriti Paul, Reza Shirkavand, Mayuka Jayawardhana, Alireza Ganjdanesh, Heng Huang, Abhinav Bhatele, Gowthami Somepalli, and Tom Goldstein.
\newblock From pixels to prose: A large dataset of dense image captions.
\newblock \emph{arXiv preprint arXiv:2406.10328}, 2024.

\bibitem[Sun et~al.(2024{\natexlab{a}})Sun, Jiang, Chen, Zhang, Peng, Luo, and Yuan]{sun2024autoregressive}
Peize Sun, Yi Jiang, Shoufa Chen, Shilong Zhang, Bingyue Peng, Ping Luo, and Zehuan Yuan.
\newblock Autoregressive model beats diffusion: Llama for scalable image generation.
\newblock \emph{arXiv preprint arXiv:2406.06525}, 2024{\natexlab{a}}.

\bibitem[Sun et~al.(2023)Sun, Yu, Cui, Zhang, Zhang, Wang, Gao, Liu, Huang, and Wang]{sun2023generative}
Quan Sun, Qiying Yu, Yufeng Cui, Fan Zhang, Xiaosong Zhang, Yueze Wang, Hongcheng Gao, Jingjing Liu, Tiejun Huang, and Xinlong Wang.
\newblock Generative pretraining in multimodality.
\newblock \emph{arXiv preprint arXiv:2307.05222}, 2023.

\bibitem[Sun et~al.(2024{\natexlab{b}})Sun, Cui, Zhang, Zhang, Yu, Wang, Rao, Liu, Huang, and Wang]{sun2024generative}
Quan Sun, Yufeng Cui, Xiaosong Zhang, Fan Zhang, Qiying Yu, Yueze Wang, Yongming Rao, Jingjing Liu, Tiejun Huang, and Xinlong Wang.
\newblock Generative multimodal models are in-context learners.
\newblock In \emph{Proceedings of the IEEE/CVF Conference on Computer Vision and Pattern Recognition}, pages 14398--14409, 2024{\natexlab{b}}.

\bibitem[Team(2024)]{team2024chameleon}
Chameleon Team.
\newblock Chameleon: Mixed-modal early-fusion foundation models.
\newblock \emph{arXiv preprint arXiv:2405.09818}, 2024.

\bibitem[Team et~al.(2023)Team, Anil, Borgeaud, Wu, Alayrac, Yu, Soricut, Schalkwyk, Dai, Hauth, et~al.]{team2023gemini}
Gemini Team, Rohan Anil, Sebastian Borgeaud, Yonghui Wu, Jean-Baptiste Alayrac, Jiahui Yu, Radu Soricut, Johan Schalkwyk, Andrew~M Dai, Anja Hauth, et~al.
\newblock Gemini: a family of highly capable multimodal models.
\newblock \emph{arXiv preprint arXiv:2312.11805}, 2023.

\bibitem[Tong et~al.(2024)Tong, Fan, Zhu, Xiong, Chen, Sinha, Rabbat, LeCun, Xie, and Liu]{tong2024metamorph}
Shengbang Tong, David Fan, Jiachen Zhu, Yunyang Xiong, Xinlei Chen, Koustuv Sinha, Michael Rabbat, Yann LeCun, Saining Xie, and Zhuang Liu.
\newblock Metamorph: Multimodal understanding and generation via instruction tuning.
\newblock \emph{arXiv preprint arXiv:2412.14164}, 2024.

\bibitem[Touvron et~al.(2023)Touvron, Martin, Stone, Albert, Almahairi, Babaei, Bashlykov, Batra, Bhargava, Bhosale, et~al.]{touvron2023llama2}
Hugo Touvron, Louis Martin, Kevin Stone, Peter Albert, Amjad Almahairi, Yasmine Babaei, Nikolay Bashlykov, Soumya Batra, Prajjwal Bhargava, Shruti Bhosale, et~al.
\newblock Llama 2: Open foundation and fine-tuned chat models.
\newblock \emph{arXiv preprint arXiv:2307.09288}, 2023.

\bibitem[Van Den~Oord et~al.(2017)Van Den~Oord, Vinyals, et~al.]{van2017neural}
Aaron Van Den~Oord, Oriol Vinyals, et~al.
\newblock Neural discrete representation learning.
\newblock \emph{Advances in neural information processing systems}, 30, 2017.

\bibitem[Wang et~al.(2024{\natexlab{a}})Wang, Lu, Yang, Huang, Han, Hou, Zhang, and Xu]{wang2024illume}
Chunwei Wang, Guansong Lu, Junwei Yang, Runhui Huang, Jianhua Han, Lu Hou, Wei Zhang, and Hang Xu.
\newblock Illume: Illuminating your llms to see, draw, and self-enhance.
\newblock \emph{arXiv preprint arXiv:2412.06673}, 2024{\natexlab{a}}.

\bibitem[Wang et~al.(2024{\natexlab{b}})Wang, Bai, Tan, Wang, Fan, Bai, Chen, Liu, Wang, Ge, et~al.]{wang2024qwen2}
Peng Wang, Shuai Bai, Sinan Tan, Shijie Wang, Zhihao Fan, Jinze Bai, Keqin Chen, Xuejing Liu, Jialin Wang, Wenbin Ge, et~al.
\newblock Qwen2-vl: Enhancing vision-language model's perception of the world at any resolution.
\newblock \emph{arXiv preprint arXiv:2409.12191}, 2024{\natexlab{b}}.

\bibitem[Wang et~al.(2024{\natexlab{c}})Wang, Zhang, Luo, Sun, Cui, Wang, Zhang, Wang, Li, Yu, et~al.]{wang2024emu3}
Xinlong Wang, Xiaosong Zhang, Zhengxiong Luo, Quan Sun, Yufeng Cui, Jinsheng Wang, Fan Zhang, Yueze Wang, Zhen Li, Qiying Yu, et~al.
\newblock Emu3: Next-token prediction is all you need.
\newblock \emph{arXiv preprint arXiv:2409.18869}, 2024{\natexlab{c}}.

\bibitem[Wu et~al.(2024{\natexlab{a}})Wu, Chen, Wu, Ma, Liu, Pan, Liu, Xie, Yu, Ruan, et~al.]{wu2024janus}
Chengyue Wu, Xiaokang Chen, Zhiyu Wu, Yiyang Ma, Xingchao Liu, Zizheng Pan, Wen Liu, Zhenda Xie, Xingkai Yu, Chong Ruan, et~al.
\newblock Janus: Decoupling visual encoding for unified multimodal understanding and generation.
\newblock \emph{arXiv preprint arXiv:2410.13848}, 2024{\natexlab{a}}.

\bibitem[Wu et~al.(2024{\natexlab{b}})Wu, Jin, Zhang, Xu, Liu, Li, and Loy]{wu2024f}
Size Wu, Sheng Jin, Wenwei Zhang, Lumin Xu, Wentao Liu, Wei Li, and Chen~Change Loy.
\newblock F-lmm: Grounding frozen large multimodal models.
\newblock \emph{arXiv preprint arXiv:2406.05821}, 2024{\natexlab{b}}.

\bibitem[Wu et~al.(2024{\natexlab{c}})Wu, Zhang, Chen, Tang, Li, Fang, Zhu, Xie, Yin, Yi, et~al.]{wu2024vila}
Yecheng Wu, Zhuoyang Zhang, Junyu Chen, Haotian Tang, Dacheng Li, Yunhao Fang, Ligeng Zhu, Enze Xie, Hongxu Yin, Li Yi, et~al.
\newblock Vila-u: a unified foundation model integrating visual understanding and generation.
\newblock \emph{arXiv preprint arXiv:2409.04429}, 2024{\natexlab{c}}.

\bibitem[Xiao et~al.(2024)Xiao, Wang, Zhou, Yuan, Xing, Yan, Wang, Huang, and Liu]{xiao2024omnigen}
Shitao Xiao, Yueze Wang, Junjie Zhou, Huaying Yuan, Xingrun Xing, Ruiran Yan, Shuting Wang, Tiejun Huang, and Zheng Liu.
\newblock Omnigen: Unified image generation.
\newblock \emph{arXiv preprint arXiv:2409.11340}, 2024.

\bibitem[Xie et~al.(2024)Xie, Mao, Bai, Zhang, Wang, Lin, Gu, Chen, Yang, and Shou]{xie2024show}
Jinheng Xie, Weijia Mao, Zechen Bai, David~Junhao Zhang, Weihao Wang, Kevin~Qinghong Lin, Yuchao Gu, Zhijie Chen, Zhenheng Yang, and Mike~Zheng Shou.
\newblock Show-o: One single transformer to unify multimodal understanding and generation.
\newblock \emph{arXiv preprint arXiv:2408.12528}, 2024.

\bibitem[Yang et~al.(2024)Yang, Yang, Zhang, Hui, Zheng, Yu, Li, Liu, Huang, Wei, et~al.]{yang2024qwen2}
An Yang, Baosong Yang, Beichen Zhang, Binyuan Hui, Bo Zheng, Bowen Yu, Chengyuan Li, Dayiheng Liu, Fei Huang, Haoran Wei, et~al.
\newblock Qwen2. 5 technical report.
\newblock \emph{arXiv preprint arXiv:2412.15115}, 2024.

\bibitem[Yu et~al.(2023)Yu, Lezama, Gundavarapu, Versari, Sohn, Minnen, Cheng, Birodkar, Gupta, Gu, et~al.]{yu2023language}
Lijun Yu, Jos{\'e} Lezama, Nitesh~B Gundavarapu, Luca Versari, Kihyuk Sohn, David Minnen, Yong Cheng, Vighnesh Birodkar, Agrim Gupta, Xiuye Gu, et~al.
\newblock Language model beats diffusion--tokenizer is key to visual generation.
\newblock \emph{arXiv preprint arXiv:2310.05737}, 2023.

\bibitem[Yue et~al.(2024)Yue, Ni, Zhang, Zheng, Liu, Zhang, Stevens, Jiang, Ren, Sun, et~al.]{yue2024mmmu}
Xiang Yue, Yuansheng Ni, Kai Zhang, Tianyu Zheng, Ruoqi Liu, Ge Zhang, Samuel Stevens, Dongfu Jiang, Weiming Ren, Yuxuan Sun, et~al.
\newblock Mmmu: A massive multi-discipline multimodal understanding and reasoning benchmark for expert agi.
\newblock In \emph{Proceedings of the IEEE/CVF Conference on Computer Vision and Pattern Recognition}, pages 9556--9567, 2024.

\bibitem[Zang et~al.(2025)Zang, Li, Han, Zhou, and Loy]{zang2025contextual}
Yuhang Zang, Wei Li, Jun Han, Kaiyang Zhou, and Chen~Change Loy.
\newblock Contextual object detection with multimodal large language models.
\newblock \emph{International Journal of Computer Vision}, 133\penalty0 (2):\penalty0 825--843, 2025.

\bibitem[Zhai et~al.(2023)Zhai, Mustafa, Kolesnikov, and Beyer]{zhai2023sigmoid}
Xiaohua Zhai, Basil Mustafa, Alexander Kolesnikov, and Lucas Beyer.
\newblock Sigmoid loss for language image pre-training.
\newblock In \emph{Proceedings of the IEEE/CVF International Conference on Computer Vision}, pages 11975--11986, 2023.

\bibitem[Zhao et~al.(2024)Zhao, Song, Wang, Feng, Ding, Sun, Xiao, and Wang]{zhao2024monoformer}
Chuyang Zhao, Yuxing Song, Wenhao Wang, Haocheng Feng, Errui Ding, Yifan Sun, Xinyan Xiao, and Jingdong Wang.
\newblock Monoformer: One transformer for both diffusion and autoregression.
\newblock \emph{arXiv preprint arXiv:2409.16280}, 2024.

\bibitem[Zhou et~al.(2024)Zhou, Yu, Babu, Tirumala, Yasunaga, Shamis, Kahn, Ma, Zettlemoyer, and Levy]{zhou2024transfusion}
Chunting Zhou, Lili Yu, Arun Babu, Kushal Tirumala, Michihiro Yasunaga, Leonid Shamis, Jacob Kahn, Xuezhe Ma, Luke Zettlemoyer, and Omer Levy.
\newblock Transfusion: Predict the next token and diffuse images with one multi-modal model.
\newblock \emph{arXiv preprint arXiv:2408.11039}, 2024.

\bibitem[Zhu et~al.(2023)Zhu, Chen, Shen, Li, and Elhoseiny]{zhu2023minigpt}
Deyao Zhu, Jun Chen, Xiaoqian Shen, Xiang Li, and Mohamed Elhoseiny.
\newblock Minigpt-4: Enhancing vision-language understanding with advanced large language models.
\newblock \emph{arXiv preprint arXiv:2304.10592}, 2023.

\bibitem[Zhuo et~al.(2024)Zhuo, Du, Xiao, Li, Liu, Huang, Liu, Zhao, Wang, Ma, et~al.]{2024lumina}
Le Zhuo, Ruoyi Du, Han Xiao, Yangguang Li, Dongyang Liu, Rongjie Huang, Wenze Liu, Lirui Zhao, Fu-Yun Wang, Zhanyu Ma, et~al.
\newblock {Lumina-Next}: Making {Lumina-T2X} stronger and faster with {Next-DiT}.
\newblock \emph{arXiv preprint arXiv:2406.18583}, 2024.

\end{thebibliography}
}

\clearpage

\newpage
\setcounter{table}{0}
\renewcommand{\thetable}{A\arabic{table}}
\setcounter{figure}{0}
\renewcommand{\thefigure}{A\arabic{figure}}
\setcounter{section}{0}
\renewcommand{\thesection}{A\arabic{section}}

\section{Appendix}

\subsection{MAR}
In this section, we provide more details of MAR~\cite{li2024autoregressive}.
\noindent\textbf{Model Details.} The MAR models in ~\cite{li2024autoregressive} follow the encoder-decoder architecture of MAE~\cite{he2022masked}, and are trained on ImageNet1K~\cite{deng2009imagenet} for image generation. Class embeddings are fed into the MAR encoder for class-conditional generation. An extra null embedding representing an empty class is also included for unconditional generation. In Harmon, we discard class embeddings and only use the null embeddings in MAR's forward pass (referred to as buffer embeddings in the main text). As a generation model, MAR follows the common practice~\cite{2022LDM, 2023SDXL, peebles2023scalable} to compress images into VAE latents before feeding to MAR' encoder. For brevity, we omit the VAE part in our illustration of Harmon.

\begin{figure*}[t]
  \centering
\includegraphics[width=0.95\textwidth]{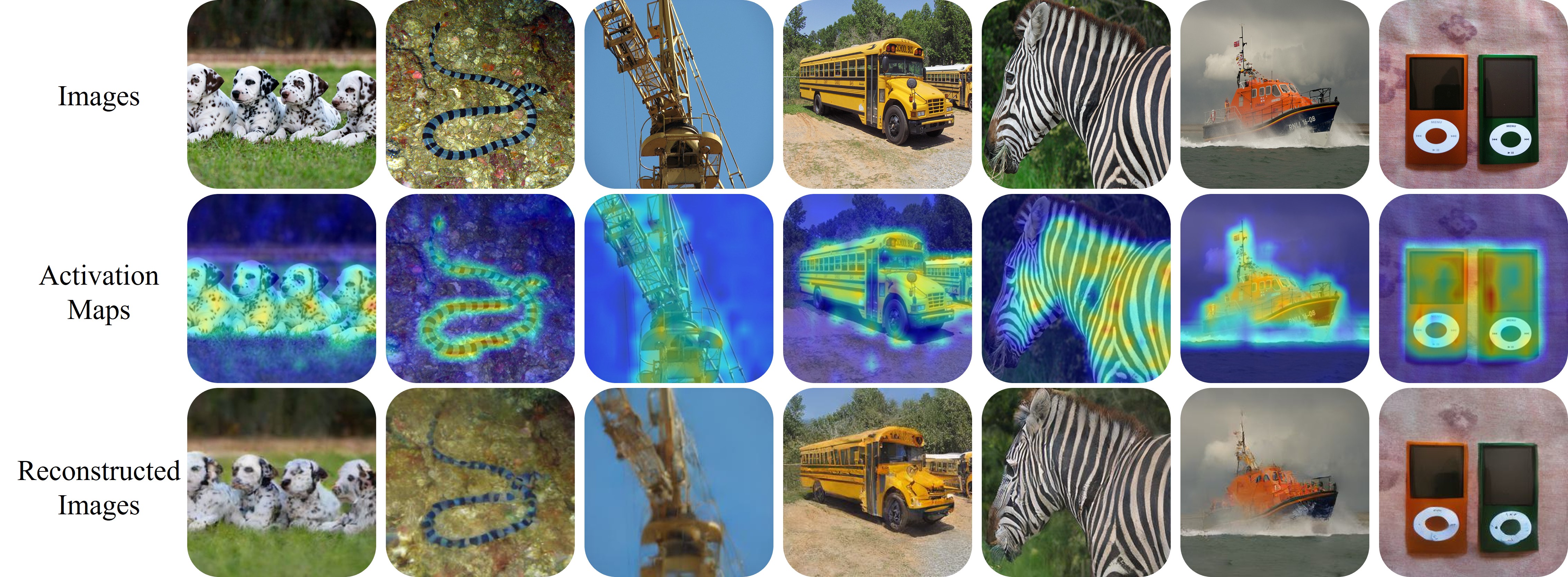}
  \caption{We visualize activations on MAR's feature maps in the second row, which reveal precise responses to visual concepts. In the third row, we observe that the features can also be mapped back to image pixels, indicating that the MAR features also comprise low-level image intrinsics.} \label{fig:mar_visualizations}
\end{figure*}

\noindent\textbf{Potential for Understanding \& Generation.} We provide more visualization results in Figure~\ref{fig:mar_visualizations} to unveil the MAR's potential for both visual understanding and generation. The feature activations in the second row of Figure~\ref{fig:mar_visualizations} indicate that the MAR encoder has grasped essential visual concepts in its generative training. Then we map the encoder features back to image contents using the MAR decoder. It is noteworthy this operation is performed in a zero-shot manner as the MAR is trained for predicting unseen patches instead of pixel-level recovery. The results in the third row of Figure~\ref{fig:mar_visualizations} suggest the MAR encoder's representation also contains intrinsic imagery features that are necessary for visual generation.

\subsection{Training Data}
We provide details of out training data, including data sources and re-captioning processes.
\subsubsection{Image Understanding}

\noindent\textbf{Stage I.} The 22M images with dense captions in stage I are sourced from LLaVA-ReCap-CC3M~\cite{li2024llava}, PixelProse~\cite{singla2024pixels}, DenseFusion~\cite{li2024densefusion} and the pre-training dataset of MiniGemini~\cite{li2024mini} and ShareGPT4V~\cite{chen2024sharegpt4v}. The dense captions in LLaVA-ReCap-CC3M are generated by LLaVA-NeXT-34B~\cite{li2024llava}. The PixelProse dataset comprises 16M images from CommonPool~\cite{gadre2023datacomp}, CC12M~\cite{changpinyo2021conceptual} and RedCaps~\cite{desai2021redcaps}, which are re-captioned by Gemini-1.0-Pro-Vision~\cite{team2023gemini}. DenseFusion labels 1M images from LAION~\cite{schuhmann2022laion} using a trained caption engine.

\noindent\textbf{Stage II.} The 20M comprehensive instruction-tuning data in stage II are from the Infinity-MM-Stage3~\cite{gu2024infinity}. And extra 5M dense-captioned images are randomly sampled from the 22M images in our stage I.

\noindent\textbf{Stage III.} In the high-quality fine-tuning stage, we directly use instruction-tuning data from LLaVA-One-Vision~\cite{li2024llava} for image understanding.

\subsubsection{Image Generation}

\noindent\textbf{Stage I.} For class-conditional image generation in stage I, we use ImageNet1K~\cite{deng2009imagenet} with 1.2M data samples, treating class names as image captions.

\noindent\textbf{Stage II.} For text-to-image generation, we first rewrite the 22M dense captions in stage I into shorter descriptions with Qwen2.5-7B-Instruct~\cite{yang2024qwen2}, using the following prompt:

\noindent``\texttt{Here is a detailed image description: <caption>. Rewrite it into a much shorter, vivid, and visually rich sentence (one or two sentences) that captures only the most essential elements and atmosphere of the scene. Ensure the description is concise, clear, and optimized for use with a text-to-image generation model.}''

\noindent Here, \texttt{<caption>} stands for the dense caption.

In addition, we import datasets specially collected for image generation, including PD12M~\cite{meyer2024public}, Megalith10M~\cite{madebyollin_megalith_10m} and LAION-Aesthetics~\cite{dclure_laion_aesthetics_12m_umap}. Like the prior 22M dense caption data, the PD12M dataset is originally labelled with detailed image descriptions. Therefore, we also use Qwen2.5-7B-Instruct to re-write all the image descriptions with the prompt defined above. For Megalith10M, we directly use the short captions provided by \cite{flickr-megalith-10m-internvl2-multi-caption}. For LAION-Aesthetics, we crawled 6M images using their urls and labelled them with precise generation prompts by Qwen2-VL-72B~\cite{wang2024qwen2}.

In total, we collect 50M data samples for training text-to-image generation in stage II.

\noindent\textbf{Stage III.} For high-quality text-to-generation, we apply an aesthetic prediction model~\cite{dclure_laion_aesthetics_12m_umap} to score the 50M images in stage II. Only images with aesthetic scores beyond 6.5 are preserved. Further, we discard images with extreme height-width ratios. Finally, 10M images are selected for stage II. Additionally, we obtain 6M synthetic images from JourneyDB~\cite{pan2023journeydb} and Text-to-Image-2M~\cite{t2i2m} to further enhance visual quality.

\subsection{Visualization}

\begin{figure*}[t]
  \centering
\includegraphics[width=0.8\textwidth]{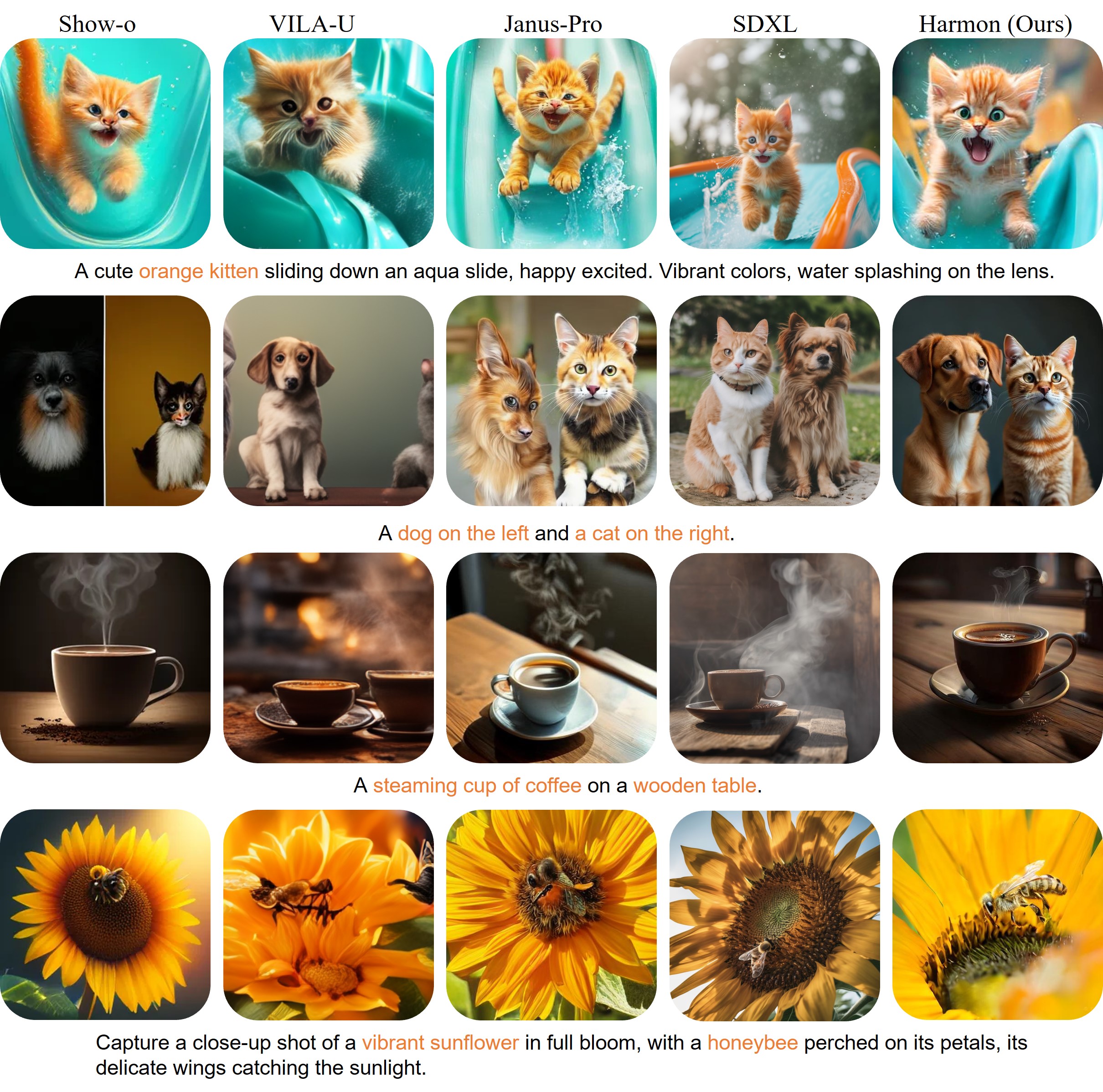}
  \caption{Qualitative comparison between Show-o-1.3B-512, VILA-U, Janus-Pro-1.5B and our Harmon-1.5B on text-to-image generation. The text below each image represents the generation prompt, with key terms guiding the generation highlighted in \textcolor{orange}{orange}. Best viewed on screen.} \label{fig:more_comparison}
\end{figure*}

\begin{figure*}[t]
  \centering
\includegraphics[width=0.8\textwidth]{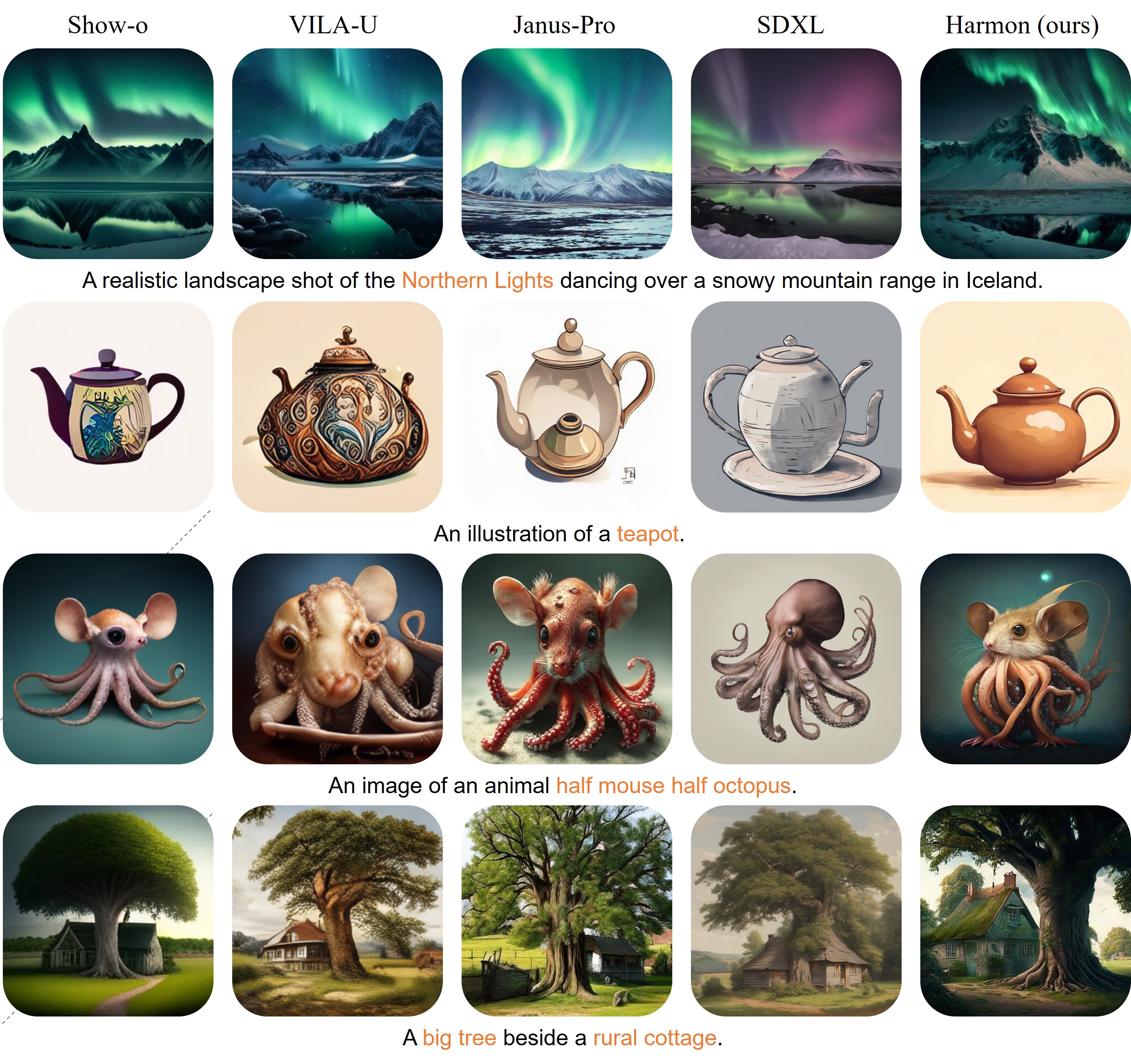}
  \caption{Qualitative comparison between Show-o, VILA-U, Janus-Pro (1.5B) and our Harmon (1.5B) on text-to-image generation. The text below each image represents the generation prompt, with key terms guiding the generation highlighted in \textcolor{orange}{orange}. Best viewed on screen.} \label{fig:more_comparison_2}
\end{figure*}

\noindent\textbf{Qualitative Comparison.}
We provide qualitative comparison on text-to-image generation in Figure~\ref{fig:more_comparison} and Figure~\ref{fig:more_comparison_2}. Here, we compare Harmon-1.5B with unified models including VILA-U~\cite{wu2024vila}, Show-o~\cite{xie2024show} and Janus-Pro~\cite{chen2025janus}(1.5B). Besides, we also include SDXL~\cite{2023SDXL}, an advanced expert model for visual generation. Harmon produces results comparable to SDXL in terms of visual quality, and exhibits better prompt-mage consistency. For example, SDXL fails to follow the position relations defined by `A dog on the left and a cat on the right' in Figure~\ref{fig:more_comparison}.

\begin{figure*}[t]
  \centering
\includegraphics[width=0.75\textwidth]{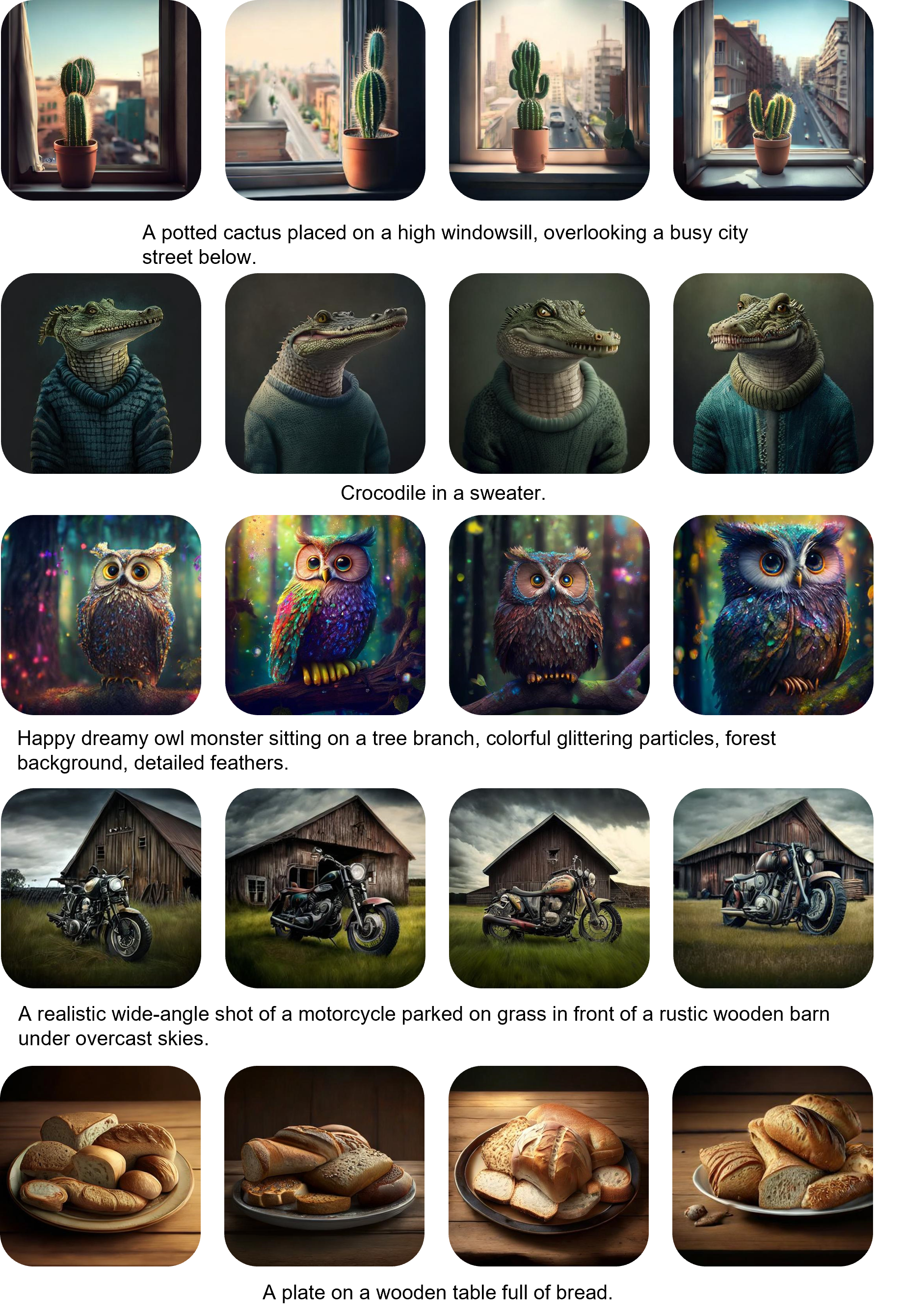}
  \vspace{-4pt}
  \caption{Text-to-image generation results by Harmon-1.5B. Our model is able to generate precise and diverse images based on text prompts.} 
\label{fig:gen_examples}
\end{figure*}

\begin{figure*}[t]
  \centering
\includegraphics[width=0.8\textwidth]{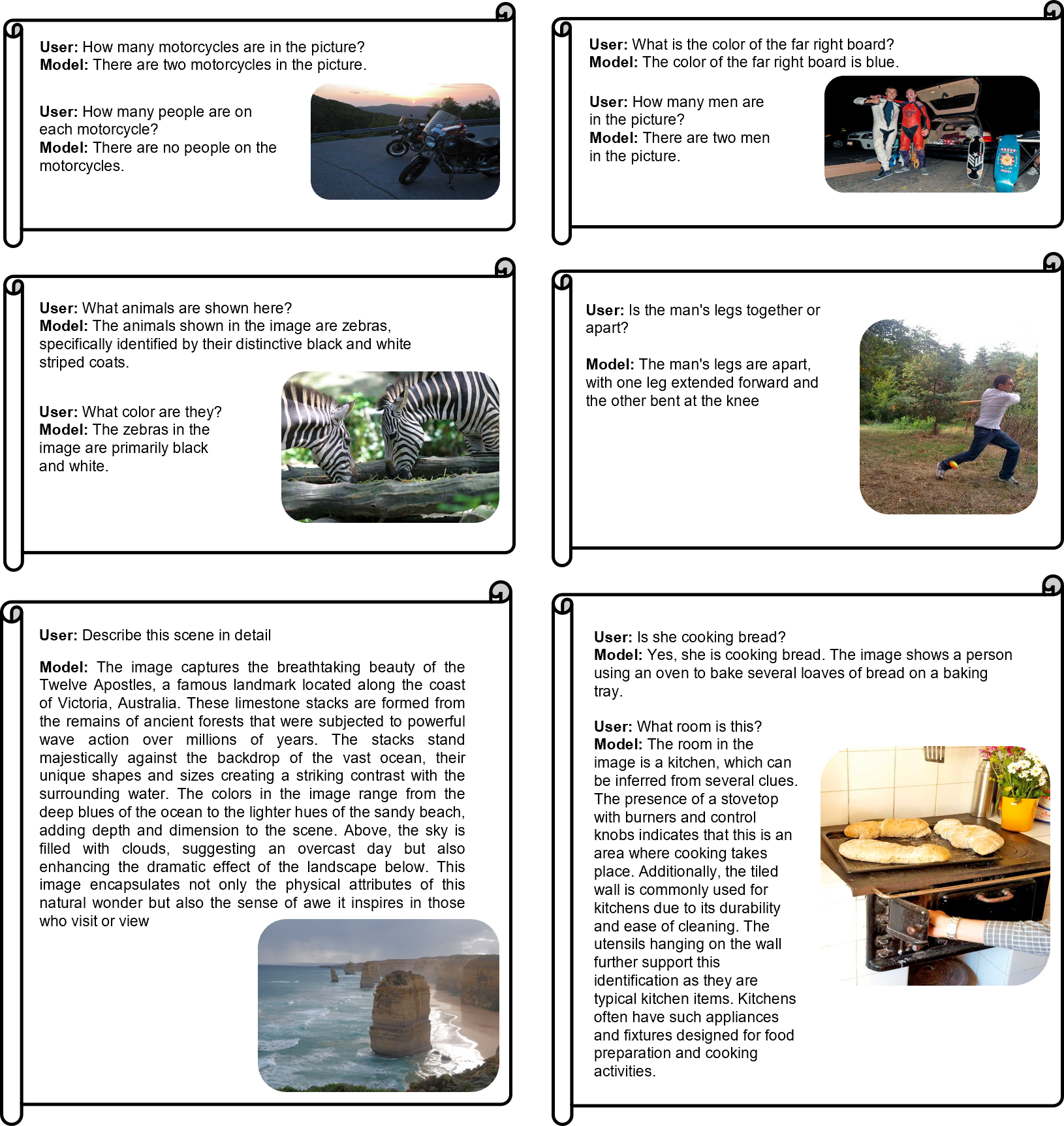}
\caption{Examples of multimodal image understanding in visual question-answering format. The results are obtained by Harmon-1.5B.} 
  \label{fig:und_examples}
\end{figure*}

\noindent\textbf{More Gen. \& Und. Results.} We show more examples of Harmon-1.5B performing text-to-image generation in Figure~\ref{fig:gen_examples} and multimodal understanding in Figure~\ref{fig:und_examples}.

\subsection{Limitations}
Despite promising results on both visual understanding and generation tasks, the current version of Harmon has the following limitations.

\noindent\textbf{Model Scale.} Our model scale is limited to 1.5B and we will further scale up the model size in the future. 

\noindent\textbf{Pre-training of MAR.} The MAR models are originally pre-trained on the 1.2M data samples of ImageNet1K, which is orders of magnitude fewer than the billion-scale training of semantic encoders like CLIP and SigLIP. This gap in data scale hinders further improvement of Harmon in understanding tasks.

\end{document}